\documentclass{article}

\usepackage{microtype}
\usepackage{graphicx}
\usepackage{subfigure}
\usepackage{nicefrac}
\usepackage{booktabs} % for professional tables
\usepackage{xcolor} % for color definitions
\usepackage{tcolorbox} % for colored boxes
\tcbuselibrary{skins} % additional library for box skins
\usepackage{FiraSans} % or any other font package of your choice
\usepackage{tikz}
\usepackage{environ}
\usepackage{wrapfig}
\usepackage{tabularx}
\usepackage{enumitem}
\usepackage{subcaption}
\usepackage{xcolor}

\usepackage[utf8]{inputenc} % allow utf-8 input
\usepackage[T1]{fontenc}    % use 8-bit T1 fonts
\usepackage{hyperref}       % hyperlinks
\hypersetup{
    colorlinks,
    linkcolor={red!50!black},
    citecolor={blue!50!black},
    urlcolor={blue!80!black}
}
\usepackage{hyperref}
\usepackage{url}
\usepackage{graphicx}

\newcommand{\red}{\textcolor[rgb]{0.8,0.0,0.0}}
\newcommand{\blue}{\textcolor[rgb]{0.0,0.0,0.8}}
\newcommand{\green}{\textcolor[rgb]{0.0,0.7,0.0}}

\usepackage{hyperref}

\usepackage[accepted]{icml2024}

\usepackage{amsmath}
\usepackage{amssymb}
\usepackage{mathtools}
\usepackage{amsthm}

\usepackage[capitalize,noabbrev]{cleveref}

\theoremstyle{plain}

\theoremstyle{definition}

\theoremstyle{remark}

\usepackage[textsize=tiny]{todonotes}
\icmltitlerunning{Understanding Stepwise Inference in Transformers}

\begin{document}

\twocolumn[
\icmltitle{Towards an Understanding of Stepwise Inference in Transformers:\\ A Synthetic Graph Navigation Model}



%\icmlsetsymbol{equal co-advising}{*}
\icmlsetsymbol{Work done as an intern at NTT}{$\dagger$}

\begin{icmlauthorlist}
\icmlauthor{Mikail Khona}{mit,ntt}
\icmlauthor{Maya Okawa}{ntt}
\icmlauthor{Jan Hula}{jan}
\icmlauthor{Rahul Ramesh}{penn,ntt}
\icmlauthor{Kento Nishi}{harvard,ntt}
\icmlauthor{Robert Dick}{michigan} 
\icmlauthor{Ekdeep Singh Lubana*}{harvard,michigan}
\icmlauthor{Hidenori Tanaka*}{ntt,harvard}
\end{icmlauthorlist}

\icmlaffiliation{penn}{University of Pennsylvania}
\icmlaffiliation{mit}{Massachusetts Institute of Technology}
\icmlaffiliation{ntt}{NTT Physics and Informatics Lab}
\icmlaffiliation{michigan}{University of Michigan}
\icmlaffiliation{jan}{CIIRC and the University of Ostrava}
\icmlaffiliation{harvard}{Center for Brain Science, Harvard University}

\icmlcorrespondingauthor{Mikail Khona}{mikail@mit.edu}
\icmlcorrespondingauthor{Ekdeep Singh}{eslubana@umich.edu}
\icmlcorrespondingauthor{Hidenori Tanaka}{hidenori.tanaka@fas.harvard.edu}
\icmlkeywords{Machine Learning, ICML}

\vskip 0.3in
]

\printAffiliationsAndNotice{\icmlEqualContribution} % otherwise use the standard text.

\begin{abstract}
Stepwise inference protocols, such as scratchpads and chain-of-thought, help language models solve complex problems by decomposing them into a sequence of simpler subproblems.
Despite the significant gain in performance achieved via these protocols, the underlying mechanisms of stepwise inference have remained elusive.
To address this, we propose to study autoregressive Transformer models on a synthetic task that embodies the multi-step nature of problems where stepwise inference is generally most useful.
Specifically, we define a graph navigation problem wherein a model is tasked with traversing a path from a start to a goal node on the graph.
Despite is simplicity, we find we can empirically reproduce and analyze several phenomena observed at scale: (i) the stepwise inference reasoning gap, the cause of which we find in the structure of the training data; (ii) a diversity-accuracy tradeoff in model generations as sampling temperature varies; (iii) a simplicity bias in the model's output; and (iv) compositional generalization and a primacy bias with in-context exemplars.
Overall, our work introduces a grounded, synthetic framework for studying stepwise inference and offers mechanistic hypotheses that can lay the foundation for a deeper understanding of this phenomenon.
\end{abstract}

\section{Introduction}
\label{sec:intro}

\begin{figure}
\centering
\includegraphics[width=0.92\linewidth]{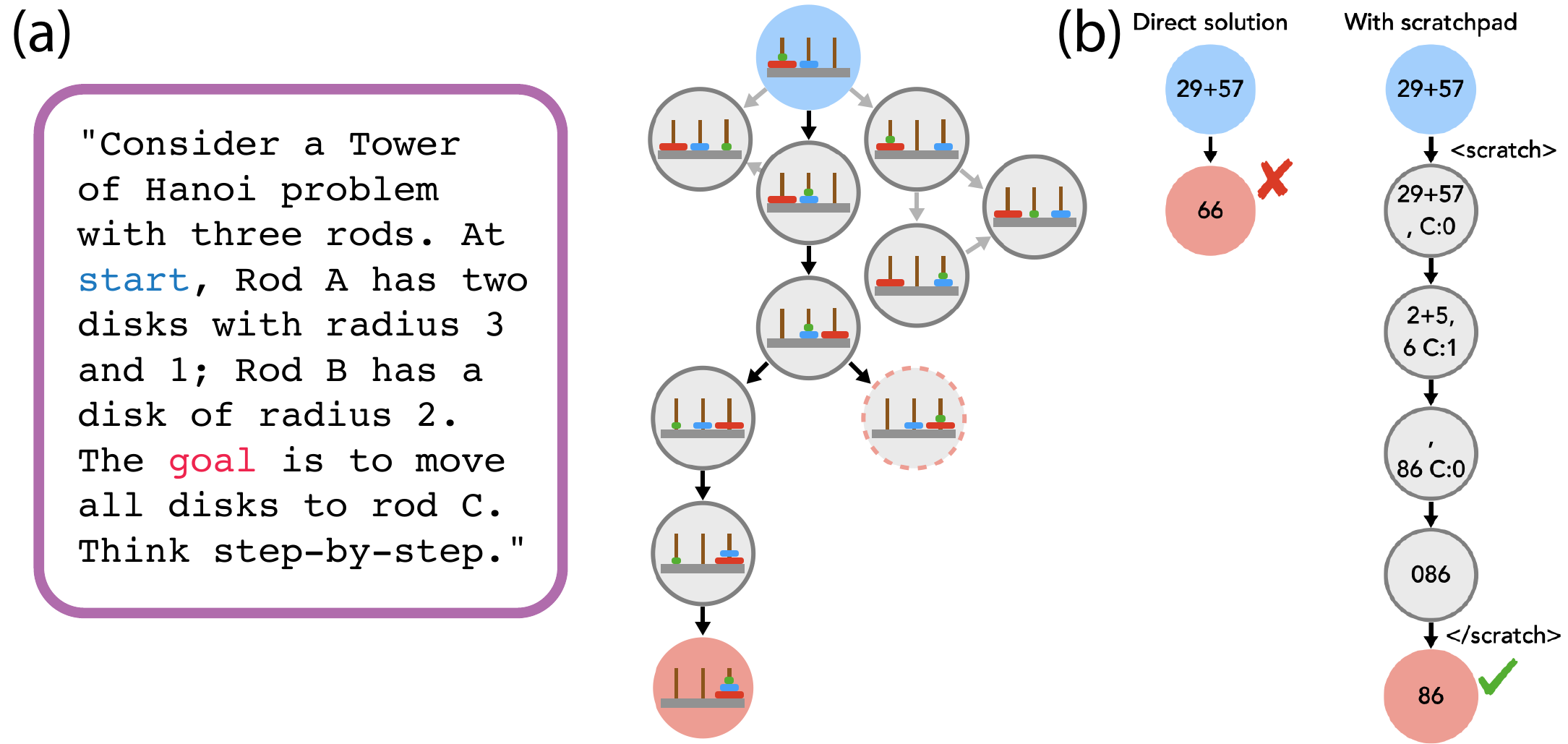}
\caption{
  \textbf{Examples of stepwise inference protocols and how they can be cast as a graph navigation problem.} 
  (a) Zero-shot chain-of-thought~\citep{kojima2022large} involves asking a model to produce intermediate outputs to perform complex multi-step computations, such as solving the Tower of Hanoi problem.
  Casting the configurations of the rods in Tower of Hanoi as nodes of a graph, we can see that the problem is essentially traversal over states describing different configurations of the setup to reach the desired configuration (the goal state).
  (b) Scratchpad~\citep{nye2021show} improves LLMs' ability to perform complex multi-step computations, such as arithmetic, when they write intermediate computation steps to a buffer called a scratchpad.
}
\label{fig:graph_navigation}
\vspace{-20pt}
\end{figure}

Transformers, the backbone of large language models (LLMs), have revolutionized several domains of machine learning~\citep{openai2023gpt4technicalreport, anil2023palm, gemini2023, touvron2023llama}. 
An intriguing capability that emerges with training of Transformers on large-scale language modeling datasets is the ability to perform \textbf{stepwise inference}, such as \textit{zero-shot chain-of-thought (CoT)}~\citep{kojima2022large}, use of \textit{scratchpads}~\citep{nye2021show}, \textit{few-shot CoT}~\citep{wei2022chain}, and variants of these protocols~\citep{creswell2022selection, yao2023tree, besta2023graph, creswell2022faithful,press2022measuring}. 
Specifically, in stepwise inference, the model is asked to or shown exemplars describing how to \textit{decompose a broader problem into multiple sub-problems}.
Solving these sub-problems in a \textit{step-by-step} manner simplifies the overall task and significantly improves performance (see Fig.~\ref{fig:graph_navigation}).
Arguably, stepwise inference protocols are the workhorse behind the ``sparks'' of intelligence demonstrated by LLMs~\citep{bubeck2023sparks,suzgun2022challenging, lu2023emergent, huang2022towards}---yet, their inner workings are poorly understood.

Motivated by the above, we aim to design and study an abstraction which enables a precise understanding of stepwise inference in Transformers.
Specifically, we argue that tasks which see maximum benefit from stepwise inference can be cast as a \textbf{graph navigation} problem: given an input describing the data to operate on and a goal to be achieved, a sequence of primitive skills (e.g., ability to perform arithmetic operations) is chained such that each skill acts on the previous skill's output, ultimately to achieve the given goal.
If the input data, the final goal, and the sequence of intermediate outputs are represented as a sequence of nodes of a graph, along with primitive skills as edges connecting these nodes, the overall task can be re-imagined as navigating nodes of the graph via the execution of primitive skills.
Several logical reasoning problems come under the purview of this abstraction~\citep{lavalle2006planning, cormen2022introduction, momennejad2023evaluating, dziri2023faith, saparov2023language}: e.g., in Fig.~\ref{fig:graph_navigation}a, we show how the problem of Tower of Hanoi can be decomposed into simpler sub-problems.
%which involve moving from the current configuration of discs (start/intermediate nodes) to the desired state (goal node). 
See also Appendix~\ref{sec:why_graph} for several more examples.

\textbf{This work.} We design a graph navigation task wherein a Transformer is trained from scratch to predict whether two nodes from a well-defined graph can be connected via a path.
A special prefix indicates to the model whether it can generate intermediate outputs to solve the task, i.e., if it can generate a sequence of nodes to infer a path connecting the two nodes; alternatively, exemplars demonstrating navigation to ``regions'' of the graph are provided.
Our framework assumes the model has perfect skills, i.e, any failures in the task are a consequence of incorrect plans for navigating the graph. 
This is justified because a skill-based failure is the most trivial mechanism via which stepwise inference protocols can fail; in contrast, inability to plan is an independent and underexplored axis for understanding stepwise inference.
% 
% We conduct evaluations to address several questions: (i) given a pair of nodes never seen in training data, can the model infer a path that connects them; (ii) given multiple paths connecting a pair of nodes, which path does the model prefer; (iii) if the training data only involves paths less than a given length, can the model learn to infer paths between nodes connected at larger lengths; (iv) how do inference hyperparameters (e.g., sampling temperature) affect the model's performance; and (v) what is the role of in-context exemplars?
% 
Overall, we make the following contributions.
\begin{itemize}[leftmargin=12pt, itemsep=1pt, topsep=-1pt, parsep=1pt, partopsep=1pt]

\item \textbf{A Framework for Investigating Stepwise Inference.} We propose a synthetic graph navigation task as an abstraction of scenarios where stepwise inference protocols help Transformers improve performance, showing that we can replicate and explain behaviors seen with use of stepwise inference in prior work. For instance, the structure of the data generating process (the graph) impacts whether stepwise inference will yield any benefits~\citep{prystawski2023think}. We identify further novel behaviors of stepwise inference as well, such as the existence of a tradeoff between diversity of outputs generated by the model and its accuracy with respect to inference hyperparameters (e.g., sampling temperature). 

\item \textbf{Demonstrating a Simplicity Bias in Stepwise Inference.} When multiple solutions are possible for an input, we demonstrate the existence of a \textit{simplicity bias}: the model prefers to follow the shortest path connecting two nodes. We assess this result mechanistically by identifying the underlying algorithm learned by the model to solve the task, showing the bias is likely a consequence of a ``pattern matching'' behavior that has been hypothesized to cause LLMs to fail in complex reasoning problems~\citep{dziri2023faith}. 
% To our knowledge, we are the first to explain this hypothesis in detail.

\item \textbf{Controllability via In-Context Exemplars.} We show the model's preferred path to navigate between two nodes can be controlled via use of in-context exemplars. We use this setup to evaluate the model's ability to generalize to paths of longer length and the influence of exemplars which conflict with each other, i.e., that steer the model along different paths.

\end{itemize}

\begin{figure*}
\centering
\includegraphics[width=0.9\linewidth]{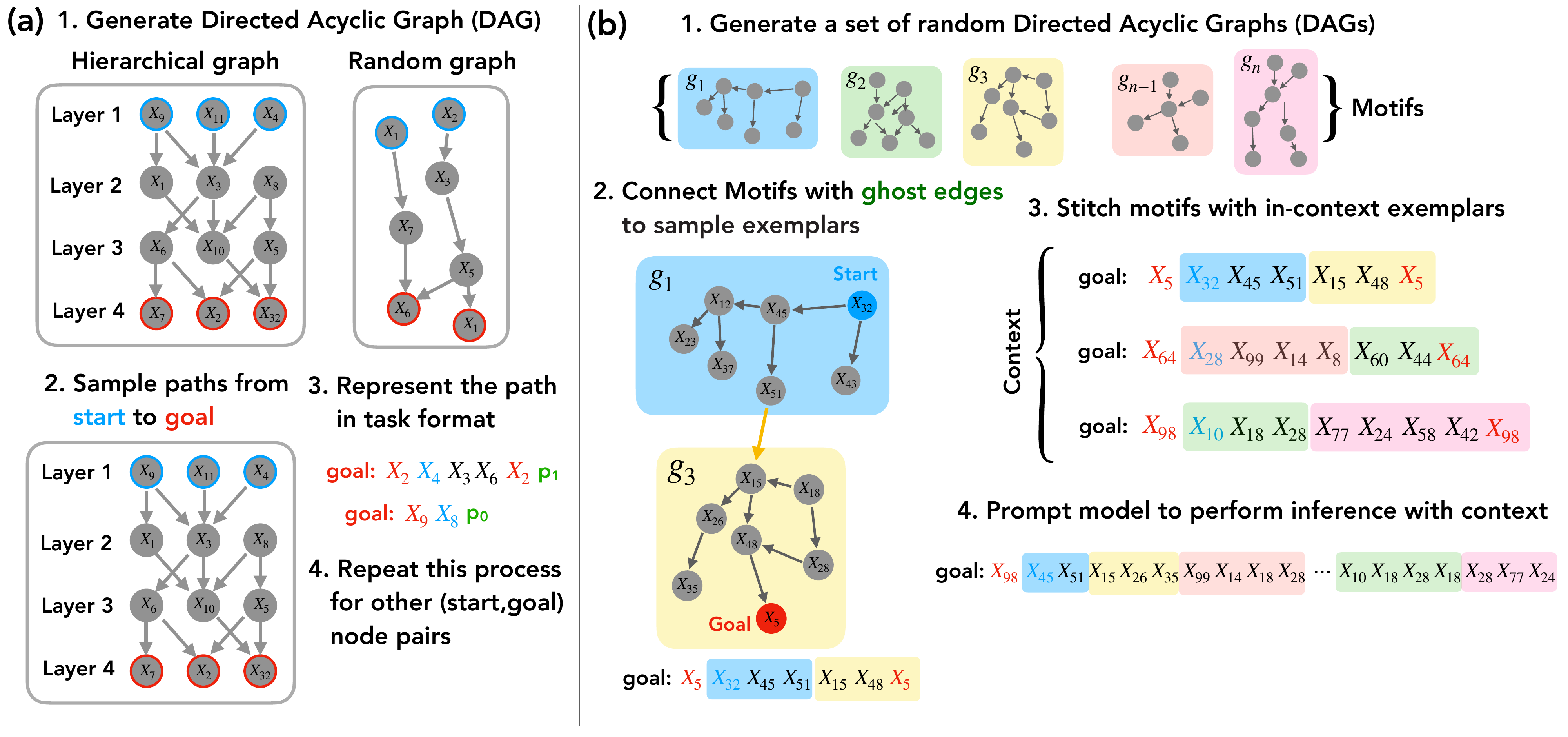}
\caption{\label{fig:dpg_both}
\textbf{Data generating process.}
(a) \textbf{In absence of exemplars.} This figure illustrates the step-by-step process of generating a training dataset using a single underlying graph. 1) A directed acyclic graph (DAG) is generated, which can be either hierarchically structured or Bernoulli. 2) A start node and a goal node are selected. 3) All possible paths connecting the start and goal nodes are sampled, and one path is randomly selected. 4) The chosen path is then represented in a task-specific format. (b) \textbf{In presence of exemplars.} The process of generating a training dataset by combining multiple subgraphs (motifs) involves the following. (1.) Start by building a set of Bernoulli directed acyclic graphs (DAGs). (2.) Pick a subset of $K$ of these DAGs $ \{g_{i_1}, g_{i_2}, .. g_{i_K} \}$ and connect them together using "ghost edges" to create a chain of motifs $ g_{i_1} \mapsto g_{i_2} \mapsto \dots \mapsto g_{i_K}$. (3.) Sample exemplars from every pair of motifs that have been connected by a ghost edge to construct the context. (4.) Now choose a start node $\blue{X_s} \in g_{i_1}$ and a goal node $\red{X_g} \in g_{i_K}$ and construct a sequence passing through the whole chain of motifs. 
}
\end{figure*}

%%%%%%%%%%%%%%%%%%%%%%%%%%%%%%%%%%%%%%%%%%%%%%%%%%%
% Setup
%%%%%%%%%%%%%%%%%%%%%%%%%%%%%%%%%%%%%%%%%%%%%%%%%%%
\section{Stepwise Inference as Graph Navigation}

In this section, we define our setup for studying how stepwise inference aids Transformers in solving complex reasoning problems.
Specifically, we define a graph navigation task wherein, given a start and a goal node, a Transformer is autoregressively trained to produce a sequence of nodes that concludes at the goal node.
In our experiments, we consider two scenarios: one where in-context exemplars are \textit{absent} (see Fig.~\ref{fig:dpg_both}a) and another where they are \textit{present} (see Fig.~\ref{fig:dpg_both}b). The former scenario emulates protocols such as the scratchpad and zero-shot Chain of Thought (CoT)~\citep{kojima2022large, nye2021show}, while the latter models few-shot CoT~\citep{wei2022chain}. 
In Section \ref{sec:dags}, we set up our experiment to explore these two scenarios. In the subsequent sections, we explicitly analyze the benefits of stepwise inference in both scenarios: without in-context exemplars (Section \ref{sec:noexemplars}) and with in-context exemplars (Section \ref{sec:exemplars}).
We refer the reader to a detailed related work on stepwise inference protocols in Appendix~\ref{sec:related_work} and further discussion on graph navigation as a model of stepwise inference which is in Appendix~\ref{sec:why_graph}.

%We divide our setup into two parts: when in-context exemplars are \textit{absent} (Fig.~\ref{fig:dpg_both}a) versus \textit{present} (Fig.~\ref{fig:dpg_both}b).
% 
%The former models protocols like scratchpad and zero-shot CoT~\citep{kojima2022large, nye2021show}, while the latter models few-shot CoT~\citep{wei2022chain}.
% 
%This division allows us to explicitly analyze the benefits of stepwise inference in the presence of extraneous context, which may influence and interact with the internalized knowledge of a model~\citep{zheng2023can} and hence its execution.
% 
% Similar to logical reasoning, the task requires the model produce a \textit{locally} correct step, i.e., each step taken by the model must be valid, and \textit{globally meaningful} sequence of steps, i.e., the sequence of steps must reach the goal node. 
% 
% This setup enables us to control various knobs: (1) the structure of the underlying graph, (2) the content of the training samples during pre-training, and (3) the information provided to the model in-context before prompting the model with the goal and start nodes. 

%%%%%%%%%%%%%%%%%%%%%%%%%%%%%%%%%%%%%%%%%%%%%%%%%%%
% DAGs
%%%%%%%%%%%%%%%%%%%%%%%%%%%%%%%%%%%%%%%%%%%%%%%%%%%
\subsection{Preliminaries: Bernoulli and Hierarchical DAGs}
\label{sec:dags}

We use \textbf{directed acyclic graphs (DAGs)} to define our graph navigation tasks. 
DAGs are a natural mathematical abstraction to study multi-step, logical reasoning problems: e.g., as discussed in~\citet{dziri2023faith}, the output of any deterministic algorithm can be represented as a DAG. 
Specifically, a DAG is defined as $G := (N, E)$, where $N := \{X_i\}_{i=1}^{|N|}$ denotes the set of nodes in the graph and $E := \{(X_i,X_j)\}_{ X_i, X_j \in N}$ denotes the set of directed edges across the nodes. 
The edges of a DAG are captured by its \textit{adjacency matrix} $A$, where $A_{ij} = 1$ if $(X_i,X_j) \in E$.
A \textit{directed simple path} is a sequence of distinct nodes of $G$ which are joined by a sequence of edges. 
If two nodes are connected via a directed simple path, we call them \textbf{path-connected}.
The first node of a path is referred to as the \blue{start node}, which we denote as $\blue{X_s}$, and the last node as the \red{goal node}, which we denote as $\red{X_g}$.

We briefly discuss the process of construction of DAGs used in our work and how paths are sampled from them; a more thorough description is provided in Appendix~\ref{sec:graph_structures}. 
We define a \textbf{Bernoulli DAG} of $N$ nodes, whose adjacency matrix has an upper triangular structure with Bernoulli entries with edge density $p$, such that $p(A_{ij} = 1) = p$. 
We ensure that all nodes have at least one edge (see Fig.~\ref{fig:dpg_both}a). The resulting DAG exhibits a bell-shaped path length distribution (see Fig.~\ref{fig:path_properties} in Appendix \ref{sec:graph_structures}).
%% We constrain the DAG to be connected (see Fig.~\ref{fig:dpg_both}a); this results in a bell-shaped path length distribution (see Fig.~\ref{fig:path_properties}). 
% 
We also define a \textbf{hierarchical DAG}, wherein the nodes follow a feedforward, layered structure such that all nodes at a given layer are only connected to nodes in the following layer (see Fig.~\ref{fig:dpg_both}a). 
In particular, for every node $n_l$ in layer $l$ and $n_{l+1}$ in layer $l+1$, we draw a directed edge $(n_l,n_{l+1})$ with probability $p$, which we refer to as \textbf{edge density}. 
On average, between any two layers of a hierarchical DAG, there are $pN^2$ edges and each node in an intermediate layer has an out-degree and in-degree of $pN$. 
The number of paths from a particular node in layer $l$ to layer $l'>l$ is exponential and given by $(pN)^{l'-l}$; this is quantified in the path length distribution shown in Appendix Fig.~\ref{fig:path_properties}. 
For both graph structures, \textbf{source nodes} are nodes that do not have any parent nodes, and the nodes that do not have any children nodes are \textbf{sink nodes}. 
%For hierarchical graphs, nodes of layer 1 are the source nodes and nodes of layer L are the sink nodes,
% A key difference between the two graph structures we use is the minimal path length: in the hierarchical DAG, intermediate layers must be traversed before reaching the goal node, resulting in longer path lengths compared to the Bernoulli DAG (see Fig.~\ref{fig:path_properties} in Appendix \ref{sec:graph_structures}). 
%% A key difference of the two graph structures we use is different minimal path length: when the DAG is hierarchical, intermediate layers must be visited before reaching the goal node,  (see Fig.~\ref{fig:path_properties}), whereas when the DAG is unstructured, there is a uniform probability that two nodes are connected and there is no explicit notion of hierarchy. 
% 
%%Accordingly, for the hierarchical structure, we can study a model's ability to perform length generalization: e.g., if only sequences with length $\leq l$ were shown in training to learn the navigation task, can the model generalize to sequences of length $>l$?
% We quantify the number of distinct paths between two randomly selected nodes and the distribution of lengths of paths in Fig.~\ref{fig:path_properties}.

%%%%%%%%%%%%%%%%%%%%%%%%%%%%%%%%%%%%%%%%%%%%%%%%%%%
% Zero shot CoT
%%%%%%%%%%%%%%%%%%%%%%%%%%%%%%%%%%%%%%%%%%%%%%%%%%%
\subsection{Modeling stepwise inference without exemplars}
\label{sec:noexemplars}

Zero-shot CoT~\citep{kojima2022large} and scratchpads~\citep{nye2021show} represent two examples of stepwise inference protocols that do not rely on exemplars. 
For instance, in the zero-shot CoT approach, the input of the model is augmented with the phrase \texttt{let's think step by step}. This encourages the model to generate outputs that elaborate on the intermediate steps required to solve the target problem, thereby enhancing accuracy by breaking down the target problem into several simpler problems.
%For example, in zero-shot CoT, one merely appends to the input ``\texttt{let's think step by step}''. 
% 
%This leads to the model producing outputs of intermediate steps involved in solving the broader problem, yielding improved accuracy via simplifcation.
%
%We note again that since no exemplars are provided, the model must rely on its internalized knowledge to infer which steps should be performed to solve the problem.
% To model the protocols above, we first define a DAG $G$ following the process discussed in Sec.~\ref{sec:dags}.

To compare the model's performance with stepwise inference and without stepwise inference (i.e., direct inference), we create two datasets: one including intermediate steps and the other without them. Each dataset is subsequently used to train distinct models. During the test phase, we present these trained models with pairs of nodes and task them to determine the existence of a path between the nodes. A model's performance is assessed based on its accuracy in classifying whether a path exists.

Fig.~\ref{fig:dpg_both}a shows how we generate the datasets above. First, we define a DAG denoted as $G$. Within this graph, for each dataset instance, we sample a start node $\blue{X_{\text{s}}}$ and a goal node $\red{X_{\text{g}}}$ and then identify all feasible paths between these two nodes. From the identified paths, we select one to form a sequence of tokens, $S$. This procedure is iterated for other node pairs within the graph $G$ to compile the complete dataset. 
For the dataset with stepwise inference, we use all the intermediate steps, including the start node \blue{$X_{\text{s}}$} and the goal node \red{$X_{\text{g}}$}, to form $S$. For the dataset without stepwise inference (i.e., direct inference), we only use the start node \blue{$X_{\text{s}}$} and the goal node \red{$X_{\text{g}}$}. 
%In stepwise inference case, the model should produce intermediate outputs; accordingly, a sequence of nodes connecting the start and goal nodes is added the $S$; 
% 
%In the other case, no further tokens are added to $S$ and path-connectedness has to be inferred directly---we call this protocol \textbf{direct inference} hereafter.
% 
We introduce a binary variable $\mathtt{\green{path}}\in\{\green{\mathtt{p_1}}, \green{\mathtt{p_0}}\}$ to denote whether there is a path between the start and goal nodes. We append the `path' token $\green{\mathtt{p_1}}$ to the end of the sequence $S$ if there is at least one path between the start and goal nodes; otherwise, we append the `no path' token $\green{\mathtt{p_0}}$. 
% We also append a binary variable $\mathtt{\green{path}}$ to $S$: if the start and goal node are connected, special token $\green{\mathtt{p_1}}$ is used to replace the variable; else $\green{\mathtt{p_0}}$ is used.

\textbf{Example:} For the example path in Fig.~\ref{fig:dpg_both}a, in the dataset with stepwise inference, the sequence of tokens $S$ includes the intermediate steps and takes the form $\red{\mathtt{goal:}X_2} \, \blue{X_4} \, X_3 \, X_6 \, \red{X_2} \, \green{\mathtt{p_1}}$.
For the dataset without stepwise inference (i.e., direct inference), the sequence $S$ does not contain intermediate steps and has the form $\red{\mathtt{goal:}X_2} \, \blue{X_4} \, \green{\mathtt{p_1}}$. 
%To promote learning to directly infer whether the start and goal nodes can be navigated between, the overall training dataset will contain samples of the form $\red{\mathtt{goal:}X_g} \blue{X_s} \, \green{\mathtt{p_1}}$.
% 
%To promote learning to produce intermediate outputs, samples of the form $\red{\mathtt{goal:}X_g} \blue{X_s} \, X_1 \dots X_k \red{X_g} \green{\mathtt{p_1}}$ are added to the dataset. Here, $X_1 \dots X_k$ is a path connecting the nodes.

%%%%%%%%%%%%%%%%%%%%%%%%%%%%%%%%%%%%%%%%%%%%%%%%%%%
% Few shot CoT
%%%%%%%%%%%%%%%%%%%%%%%%%%%%%%%%%%%%%%%%%%%%%%%%%%%
\subsection{Modeling stepwise inference with exemplars}
\label{sec:exemplars}
Here we examine the influence of stepwise inference on model performance when in-context exemplars are present. This scenario is prominently exemplified by protocols based on few-shot CoT~\citep{wei2022chain, creswell2022selection}.
%: e.g. do they override the model's internalized knowledge to alter the steps it would have taken in the absence of exemplars? Our setup is designed to address such questions (see Fig.~\ref{fig:dpg_both}b). 
% We modify the process from Sec.~\ref{sec:noexemplars} , introduce to now include a \textit{set of DAGs} that we refer to as \textbf{motifs}, denoted by $\widehat{g} = \{g_i\}_{i=1}^{K}$. 

Specifically, we extend the setup with a single DAG described in Section~\ref{sec:noexemplars} by incorporating a \textit{set of DAGs}, which we call \textbf{motifs}. The data generation process is shown in Fig.~\ref{fig:dpg_both}b. 
First, we generate a set of $n$ Bernoulli DAGs denoted by $\widehat{g} = \{g_i\}_{i=1}^{n}$ and randomly select a subset of $K$ motifs from this set $\{g_{j_1}, g_{j_2}, \dots, g_{j_K}\} \subset \widehat{g}$. Then, we add edges between the sink node of each motif $g_{j_k}$ and the source node of the subsequent motif $g_{j_{k+1}}$, forming a chain of motifs $g_{i_1} \mapsto g_{i_2} \mapsto \dots \mapsto g_{i_K}$. These interconnecting edges are termed \textbf{ghost edges}.
We sample paths from each pair of motifs linked by a ghost edge to establish the context. We select a start node from the sink nodes of one motif, $\blue{X_s} \in g$, and a goal node from the source nodes of a different motif, $\red{X_g} \in g'$, then sample a path between them, denoted as $e_{gg'}$. This procedure generates a sequence of nodes spanning across motifs, $g \rightarrow g'$, including exactly one ghost edge. We refer to this as an \textbf{exemplar sequence} and use them as in-context samples. 
%A sequence is defined by sampling paths from a graph that is a composition of a randomly selected subset of these motifs, denoted as $G := f \left(g_{i_1}, g_{i_2}, \dots g_{i_K}\right)$, where all motifs $g_{i_j} \in \widehat{g}$ and $f$ is a random process that defines special edges called \textbf{ghost edges} to connect motifs and define the graph; in particular, ghost edges connect a sink node of a motif $g$ with a source node of another motif $g'$.
% 
%We denote the subgraph corresponding to two motifs connected via at least one ghost edge in a graph $G$ as $g \rightarrow g'$.
% 
% We also define a notion of a \textbf{exemplar sequence}, i.e., a sequence of nodes across two motifs $g$ and $g'$ with a start node in $g$ and goal node in $g'$; this sequence is constrained to contain exactly one ghost edge (see Fig.~\ref{fig:dpg_both}b). We refer the reader to Appendix~\ref{Motif_construction} for further details.
% 
Exemplars to model few-shot CoT are represented as $e_{gg'}$ and denote a exemplar sequence from the motif $g \rightarrow g'$.
% 
% Given a start node from a motif $\blue{X_s} \in g$ and a goal node from the same or different motif $\red{X_g} \in g'$, 
Finally, we select a start node $\blue{X_s} \in g_{i_1}$ and a goal node $\red{X_g} \in g_{i_K}$. We then prompt the model to either directly output a path that connects the node pair $\blue{X_s}$ and $\red{X_g}$, or to provide exemplars demonstrating traversal between motifs within the specified context. 
% Since a graph is defined via combinations of motifs, we intentionally leave out 20\% combinations from the training data.
Recall that our graph is constructed from a combination of $K$ motifs. For the training dataset, we intentionally exclude 20\% of the combinations. For the test dataset, we randomly select motifs from the remaining combinations in $\widehat{g}$, and sample sequences that illustrate how to navigate between two nodes within this graph.
% Every time a new input is to be designed, we will randomly select motifs from the remaining combinations from $\widehat{g}$ to design a graph $G$, sampling sequences showing how to travel between two nodes from this graph. 
% 
From training data, a model can learn the structure and interconnections of motifs; yet, during testing, it faces unseen combinations of these motifs. 
% Though a model can learn the structure of the motifs themselves and how some of these motifs are connected, the graphs at test time will involve combinations of motifs that were never seen in training. 
% 
Correspondingly, \textit{the model must use the context to infer the overall structure of the graph}.
In essence, an exemplar tells the model which motifs are connected via ghost edges and hence can be navigated between.
% 
% Since different inputs have different motifs connected together, problem is often underspecified and hence it is important to infer from context which motifs are connected.

\textbf{Example:} We directly study the path of navigation outputted by the model in this setup, i.e., no special tokens are used. 
A sample is constructed by selecting motifs to define in-context exemplars, say $g_{i_1}, g_{i_2}, g_{i_3}$. 
For every successive pair of motifs, we construct an exemplar and put them together to create the context. 
To do this, we select two (start, goal) pairs: $\blue{X_{s_1}} \in g_{i_1}$, $\red{X_{g_1}} \in g_{i_2}$ and $\blue{X_{s_2}} \in g_{i_2}$, $\red{X_{g_2}} \in g_{i_3}$. 
We sample exemplar sequences starting and ending at these node pairs: one sequence from $g_{i_1}$ to $g_{i_2}$, $\red{\mathtt{goal:}X_{g_1}} \blue{X_{s_1}} X_1 \dots X_{k_1} \red{X_{g_1}}$, and another from $g_{i_2}$ to $g_{i_3}$, $\red{\mathtt{goal:}X_{g_2}} \blue{X_{s_2}}  X'_1 \dots X'_{k_2} \red{X_{g_2}}$. 
These sequences act as exemplars to be provided in context to the model when it is shown an input. 
The number of exemplars can vary from two to four, which correspond to chains of motifs of length three to five.
The input itself is defined by choosing a goal node $\red{X_{g}} \in g_{i_3}$, a start node $\blue{X_{s}} \in g_{i_1}$, and a path through an intermediate node $X_{\text{inter}} \in g_{i_2}$; e.g., $\red{\mathtt{goal:}X_{g}} \blue{X_{s}} X''_1 \dots X_{\text{inter}} \dots X''_{k_1} \red{X_{g_3}}$. 
Here, $\blue{X_{s}} X''_1 \dots X_{\text{inter}}$ is a path between motifs $g_{i_1}$ and $g_{i_2}$, while $X_{\text{inter}} \dots X''_k \red{X_{g_3}}$ is a path between motifs $g_{i_2}$ and $g_{i_3}$.
When exemplars are not provided, the model must rely on its internalized knowledge to infer whether there exist two connected motifs that can be used to move from the start to goal node.
The context exemplars simplify the problem by telling the model the motifs above are connected. 
% 
% Refer to Fig.~\ref{fig:dpg_both}b for a visualization of the process above.

%%%%%%%%%%%%%%%%%%%%%%%%%%%%%%%%%%%%%%%%%%%%%%%%%%%
% Results
%%%%%%%%%%%%%%%%%%%%%%%%%%%%%%%%%%%%%%%%%%%%%%%%%%%
\section{Results: Stepwise Navigation}
In this section, we discuss findings on how stepwise inference affects the model's ability to solve problems. We investigate two scenarios: in the absence of in-context exemplars (Section \ref{sec:result_noexemplars}) and in the presence of them (Section \ref{sec:multi_graph}). 
% We again break the section into two parts: in absence and presence of in-context exemplars.
% 
For all experiments, unless stated otherwise, we use a 2-layer Transformer defined by \citet{karpathy} to mimic the GPT architecture~\citep{brown2020language}.
For more details on the experimental setup, please refer to Appendix~\ref{sec:arch} for model architecture details and Appendix~\ref{sec:training_protocol} for training data generation and train/test split.

% \todo{Improve this list of contributions to be maximally accurate, while laying emphasis on what we show.}
% Broadly, we demonstrate that in the former case, depending on the structure of the underlying graph, there is a large gap in model performance whether stepwise inference is used or not. 
% % 
% Moreover, the model has a preference for solving the task by using the shortest path possible, which we explain via a mechanistic decomposition of the model.
% % 
% Finally, we show that in-context exemplars enable steering the model's preferred navigation paths.

%%%%%%%%%%%%%%%%%%%%%%%%%%%%%%%%%%%%%%%%%%%%%%%%%%%
% Zero-shot CoT
%%%%%%%%%%%%%%%%%%%%%%%%%%%%%%%%%%%%%%%%%%%%%%%%%%%
% \setlength{\parskip}{0pt}
\subsection{Navigation without exemplars}
\label{sec:result_noexemplars}

We assess the performance of the model by evaluating its ability to classify whether there is a path given a pair of nodes during the test phase. 
% We first assess how stepwise inference without exemplars aids the model in evaluating whether a pair of nodes is connected via a path.
% 
Specifically, we randomly sample pairs of start and goal nodes that were not seen in the training data and observe whether the model outputs either the `path' token $\green{\mathtt{p_1}}$ or the `no path' token $\green{\mathtt{p_0}}$. 

\subsubsection{Stepwise inference gap}
\begin{figure}
    \includegraphics[width=\linewidth]{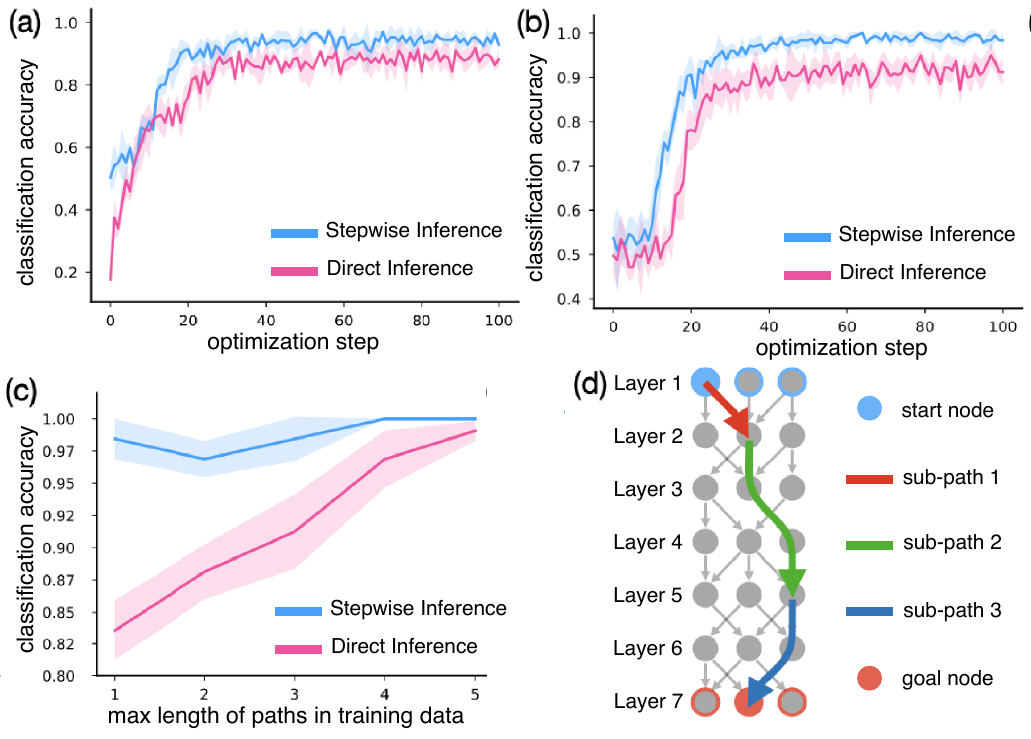}
      \caption{
  \textbf{Advantage of stepwise inference in graph navigation tasks and stitching:} 
    %(a) In the Bernoulli DAG, stepwise inference shows an advantage over direct inference in predicting whether two nodes are connected.
    %(b) This advantage is further pronounced in hierarchical graphs, where the distances between nodes can be much larger than Bernoulli graphs. 
    %(c) Probing further, we find the stepwise inference gap arises when the training set contains paths that are shorter than the paths required to connect nodes at test time. 
    %(d) This indicates that stepwise inference is beneficial when a model must trim and connect paths it has learned during training to generalize effectively: The red, green, and blue paths are \textit{subsets} of paths seen during pretraining while the combined path is one produced by the model during evaluation.
    (a) In the Bernoulli DAG, stepwise inference demonstrates an advantage over direct inference in predicting whether given node pairs are connected.
    (b) This advantage is further pronounced in hierarchical DAGs, where the distances between nodes are greater than in Bernoulli DAGs.
    (c) The stepwise inference gap arises when the training set contains paths that are shorter than the paths required to connect nodes at test time.
    (d) The stepwise inference is beneficial when the model must connect paths seen during training: the red, green, and blue paths represent subsets of paths seen during training; we find the model produces paths that combine these subsets during the test phase. 
    }
    % \vspace{-20pt}
\label{fig:stepwise_gap}
\end{figure}

%, by setting the special token $\steps{\mathtt{C}}$, elicit either stepwise or direct inference processes.
% 
% The special token $\green{\mathtt{path}}$ outputted by the model indicates whether the model expects the nodes to be path-connected\footnote{Note that technically it is possible the model deems two nodes to be connected, but produces a path that does not connect them---an unfaithful reasoning~\citep{turpin2023language}. However, we did not see this failure mode in the trained models and hence directly perform our evaluation on the output token's value itself.}.
% 
% The model's prediction is deemed accurate if it matches the ground truth, i.e., whether the nodes are actually connected in the graph.
% 
Fig.~\ref{fig:stepwise_gap} shows the accuracy of classifying `path' or `no path' for two different types of graphs: a Bernoulli graph and a hierarchical graph. 
We observe that for both types of graphs, the use of stepwise inference significantly improves the model's performance compared to direct inference, with more pronounced improvements noted for the hierarchical graph. Following \citet{prystawski2023think}, we refer to the improvement in performance observed between stepwise inference and direct inference as the ``stepwise inference gap''.  
We even simulate the effect of noisy real-world labels by introducing random corruptions into the tokens and found that the results above continue to hold, as detailed in Appendix Fig.~\ref{fig:noise_corrupted_gap}. 
\begin{figure}
\centering
\includegraphics[width=0.9\linewidth]{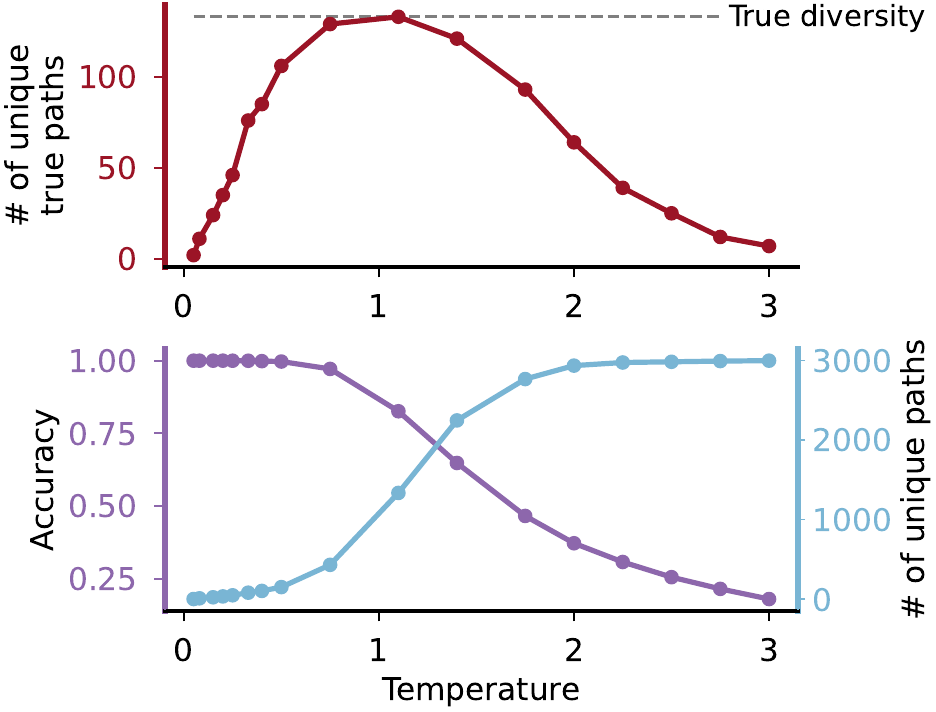}
\caption{
% \textbf{A diversity vs.\ accuracy trade-off in finite temperature stepwise inference for Transformers:} As sampling temperature is increased, the diversity of paths generated by the model from a single $(\blue{X_s},\red{X_g})$ pair increases, while the accuracy of the path decreases. This tradeoff is captured by measuring the number of unique \textit{valid} paths which is non-monotonic (top), showing the existence of an optimal temperature for sampling. The dashed line denotes the ground truth path diversity of $(\blue{X_s},\red{X_g})$. 
\textbf{Diversity vs. accuracy trade-off for different sampling temperatures of the Transformer model:} As the sampling temperature increases, the diversity of paths generated by the model also increases, while the accuracy decreases. This tradeoff is captured by measuring the number of unique \textit{valid} paths (top panel), indicating that there is an optimal temperature for sampling. The dashed line represents the ground truth path diversity. 
} \vspace{-10pt}\label{fig:temperature_tradeoff}
\end{figure}

To further probe the results above, we control for path lengths in the hierarchical graph.
Specifically, to set the maximum path length in the training data to $\Delta$, we choose a starting layer $l$ and a goal layer $l'$ such that $l' - l < \Delta$. Then, we sample starting nodes from layer $l$ and goal nodes from layer $l'$. For the test data, we select node pairs with $l' - l \geq \Delta$.
% the training data now contains start nodes from layer $l$ and goal nodes from layer $l'$ such that $l'-l<\Delta$, where $\Delta$ denotes the maximum length of paths included in the training data. 
% 
% During evaluation, we choose node pairs such that $l'-l \geq \Delta$. 
% 
% Since the number of directed simple paths for a hierarchical graph increase exponentially with path length (number of paths $\approx (pN)^{l'-l})$), for smaller values of $\Delta$, the model has observed a much smaller number of paths; see also Appendix Fig.\ref{fig:path_properties}). 
% Thus, the model must combine and piece together several different paths seen over pretraining to effectively solve the task. 
%This occurs because when the training data only includes short paths, the model needs to more effectively 'stitch' the paths observed during training, which makes direct inference more challenging. In contrast, stepwise inference maintains high accuracy across all cases. 
Results are shown in Fig.~\ref{fig:stepwise_gap}(c). We plot the classification accuracy across various values of $\Delta$ and
observe that the smaller the value of $\Delta$, the greater the stepwise inference gap becomes. 
We hypothesize this happens because when the training data only includes short paths, the model needs to more effectively `stitch' the paths observed during training, which, as a recursive task, is more feasible via stepwise inference.
% 
% This is because the shorter the paths seen during training, the more recombination the model has to do.
% 
% We hypothesize that this is where intermediate steps and stepwise inference will most improve accuracy---when the model must `stitch' together subsets of paths seen over training in flexible ways to generalize.

\subsubsection{Diversity-accuracy tradeoff with higher sampling temperatures}
%Autoregressive models rely on sampling for next-token generation. 
% 
%At low temperatures, this process is deterministic and equivalent to taking most likely token at that step, i.e., the maximum likelihood estimate. 
% 
%To address this and get a rich variety of responses, higher sampling temperatures are often used; however, the model is more likely to make mistakes by `hallucinating' plausible, but possibly incorrect, outputs at higher temperatures. 
% In our experiments above, we deterministically (zero-temperature) sampled the path for any given provided pair of start and goal nodes ($\blue{X_{s}}$, $\red{X_{g}}$) from the model. 
% 
% However, in the underlying graph, there are typically several paths connecting a pair of nodes (see Appendix~\ref{sec:graph_properties}).
% 
Here, we investigate how the sampling temperature of the autoregressive Transformer affects the diversity of the generations produced by the model and its accuracy.
To this end, we fixed the start and goal nodes and prompted the model 3,000 times, varying the sampling temperatures from 0.0 to 3.0. We define accuracy as the probability that a generated path consists of valid edges and correctly terminates at the designated goal node. Diversity is defined as the number of unique paths generated. 
As shown in Fig.~\ref{fig:temperature_tradeoff}, there is a clear trade-off between the diversity of the paths generated by the model and their accuracy. We term this phenomenon the \textbf{\textit{diversity-accuracy tradeoff}}: at lower sampling temperatures, the model generates fewer but more accurate and valid paths; in constrast, higher sampling temperatures result in greater path diversity but reduced accuracy.
Our result provides the first explicit demonstration of a trade-off between the accuracy and diversity of Transformer outputs. 
% Our results offer the first explicit demonstration of the trade-off between accuracy and diversity in Transformer outputs, marking a significant step towards a deeper understanding of this phenomenon. 
%Such understanding is crucial for generating a rich variety of responses while minimizing the risk of `hallucinating' plausible but potentially incorrect outputs. 
% 
To the best of our knowledge, this phenomena has not been quantitatively studied before.

\begin{figure}
\centering
\includegraphics[width=0.9\linewidth]{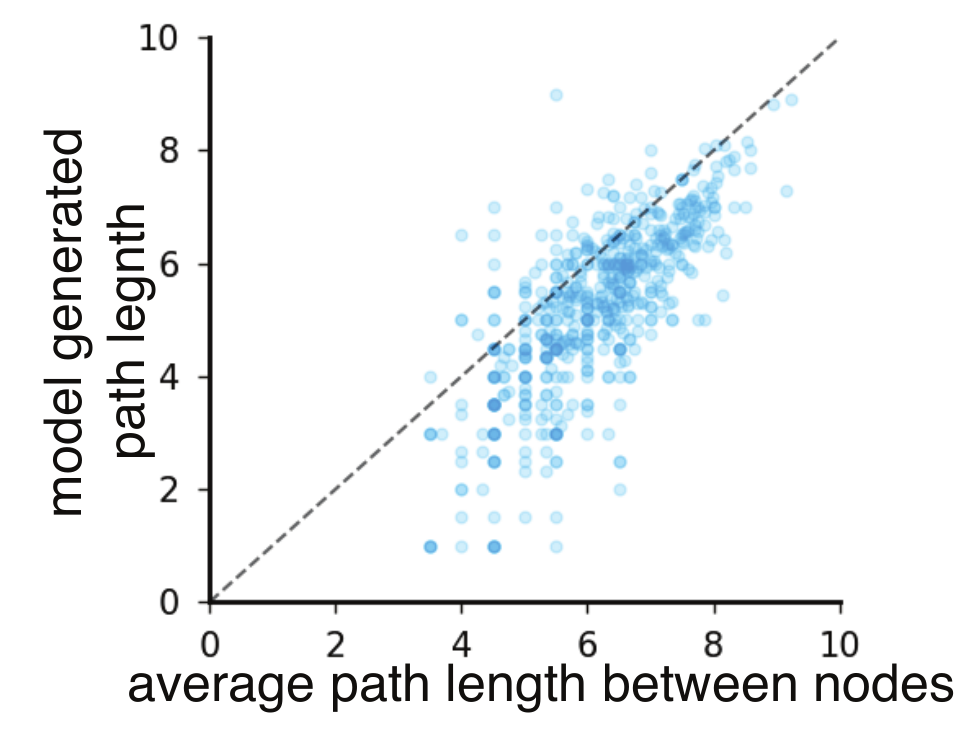}
\caption{\textbf{Model outputs are biased toward shorter paths.} We compared the average lengths of ground-truth paths for a specific set of node pairs and the paths produced by the model for these same pairs in the Bernoulli DAG. We observe that the model tends to generate shorter paths than the actual ones. This observation points to a ``simplicity bias'' in the trained model towards favoring shorter over potentially more accurate or realistic paths. 
}
\label{fig:bias}
% \vspace{-18pt}
\end{figure}

\subsubsection{Preference for shorter paths}
\label{subsec:short_path_bias}
Note that there are multiple possible paths the model can choose from in the pursuit of inferring a path that connects a start and goal node. 
We showed that by increasing the sampling temperature, a diverse set of paths can be generated; however, by default, which path does the model prefer?
To evaluate this, we compare the actual path lengths between nodes in the test data with those generated by our trained model in the Bernoulli graph setup. In Fig.~\ref{fig:bias}a, we observe that the model consistently produces paths that are shorter, on average, than the paths in the ground truth DAG. This observation suggests that the model exhibits a \textit{simplicity bias}, tending to find the quickest path to solve the target problem. 
However, simplicity biases have been shown to yield oversimplification of a problem, forcing a model to learn spurious features~\citep{shah2020pitfalls, lubana2023mechanistic}.
In the context of stepwise inference, this can amount to omission of important intermediate steps, similar to `shortcut solutions' arising from pattern-matching behaviors discussed in prior work on Transformers~\citep{liu2022Transformers, dziri2023faith}.

\subsubsection{Evolution of failures in stepwise inference over training}
\begin{figure}
\centering
\vspace{-5pt}
\includegraphics[width=0.9\linewidth]{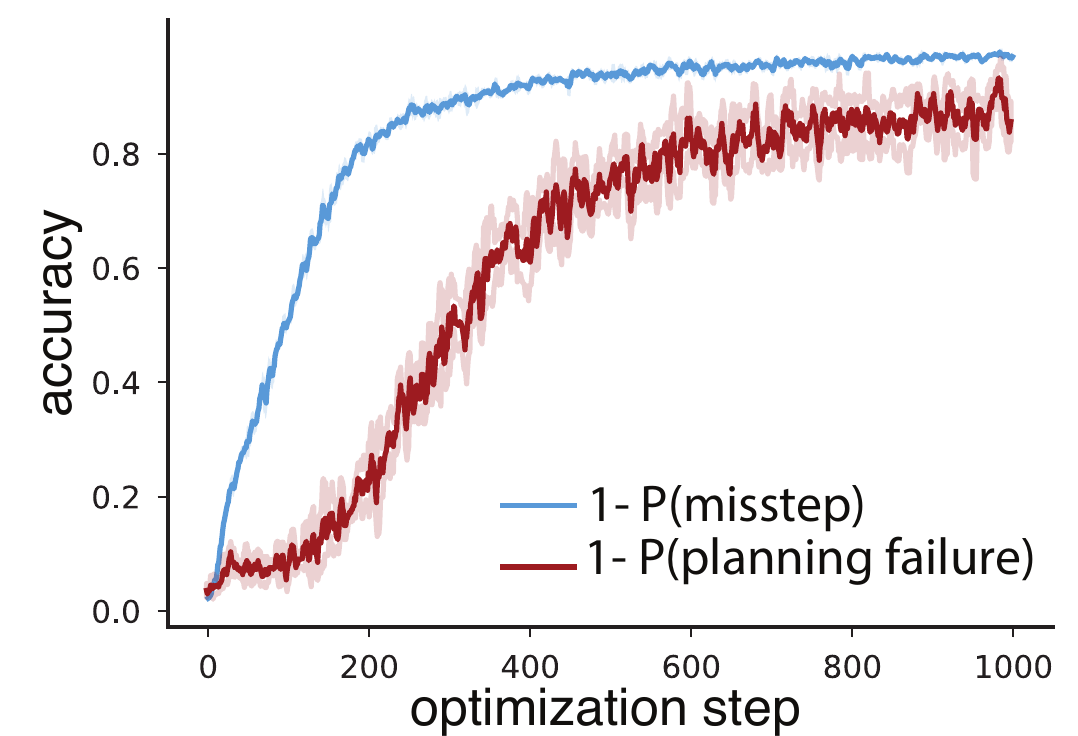}
\caption{\textbf{Learning dynamics for two failure modes: misstep and planning failure. } We measure the probability of missteps and planning failures in the model's outputs. A misstep refers to an instance where the model generates an edge that is not present in the graph, while a planning failure means that the model outputs a path that fails to reach the intended goal node. Initially, the model learns to avoid missteps. Subsequently, around the $200^{\text{th}}$ optimization step, it begins to effectively learn planning. The accuracy curves are averaged over three models, each trained with a distinct random seed. 
% \textbf{The learning dynamics of failure mode probabilities over training:} Over pretraining on a Bernoulli graph $G$, we measure the probability of generating missteps, which are defined by $(X_k, X_{k+1}) \notin G$ and the probabilities that the model-generated paths do not end at the goal $\red{X_{g}}$ specified in the prompt. The model first learns to produce correct edges (effectively learn correct bigram statistics) and then (around iteration 200 in the plot) learns the global objective of producing a path that ends at the cued goal node $\red{X_g}$. Accuracy curves are averaged over 3 trained models with different random seed. 
}
\vspace{-8pt}
\label{fig:failure}
\end{figure}
In the above discussion, we evaluated how stepwise inference assists a model in successfully completing a complex, multi-step task. We now assess how it fails.
Specifically, assume that for a given graph $G$, the model produces a sequence of nodes $\blue{X_{s}} X_1 \dots X_k \dots X_{t}$ starting at the start node \blue{$X_s$}. 
Following~\cite{saparov2023language, momennejad2023evaluating}, we define two categories of potential failures.
\begin{itemize}[leftmargin=8pt, itemsep=1pt, topsep=-1pt, parsep=-1pt, partopsep=-1pt]
\item \textbf{Misstep} $(X_k, X_{k+1}) \notin G$: An edge produced by the model does not exist in the DAG, commonly referred to as ``hallucinations''. 
\item \textbf{Planning failure} $X_t \neq \textcolor{red}{X_{g}}$: The model produces a path that does not terminate at the goal node. 
\end{itemize}

\label{sec:mechinterp}
\begin{figure*}[t]
  % \vspace{-10pt}
  \centering
    \includegraphics[width=0.96\linewidth]{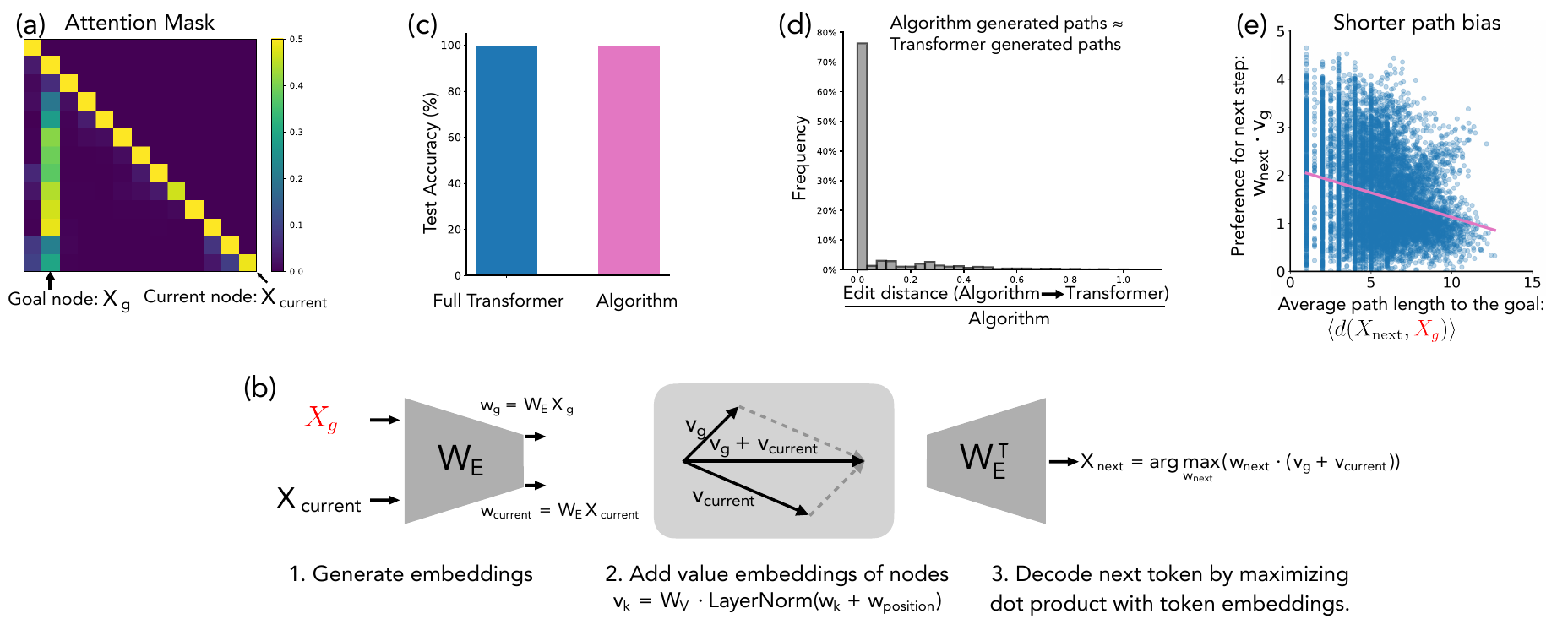}
  \caption{\textbf{Mechanistic analysis of the graph navigation algorithm: Emergent linear representation.} (a) Attention maps from the 1-layer, attention-only Transformer, highlighting the model's attention on the goal token $\textcolor{red}{X_g}$ and the current token $X_{\text{current}}$. (b) Steps of our simplified algorithm that emulates the 1-layer, attention-only Transformer are as follows. 
  (1.) We extract value embeddings for $\textcolor{red}{X_g}$ and $X_{\text{current}}$, ignoring other tokens for simplicity.
  %using the embedding matrix $W_E$, yielding $w_{g}$ and $w_{\text{current}}$.
  (2.) Next, we compute the value embeddings of the goal $v_{g}$ and current $v_{\text{current}}$ nodes and add them together $v = v_{g} + v_{\text{current}}$.
  %we apply layer normalization and weight matrix $W_V$ in the self-attention layer, yielding $v_{g}$ and $v_{\text{current}}$. Then, we take summation of $v_{g}$ and $v_{\text{current}}$ to obtain $v = v_{g} + v_{\text{current}}$. 
  (3.) We then compute the token embedding with the highest inner product with $v$, approximating the token that receives the highest logit score after the single forward pass. 
  (c) Comparison of the model's accuracy on a set of 500 held-out node pairs ($\textcolor{blue}{X_s}$, $\textcolor{red}{X_g}$) using our simplified algorithm (99.8\%) versus the full trained model (99.6\%). (d) The paths generated by the simplified algorithm almost exactly match the paths generated by the full trained model. Path similarity on 2000 held-out node pairs was compared by measuring the Levenshtein edit distance ~\citep{navarro2001guided} between paths generated by the full trained model and the simplified algorithm for the same $(\blue{X_s},\red{X_g})$ pairs. (e) The short path bias can be attributed to the inner products between the token embedding of the next chosen token $X_{\text{next}}$ and the value embedding of $\textcolor{red}{X_g}$ and $v_{g}$. We observe that nodes $X_{\text{next}}$ further away from $\textcolor{red}{X_g}$ have a lower inner product, indicating that the model's embedding of nodes reflects the underlying graph topology. The red line denotes the best least squares fit and has a slope of $-0.106$.
\vspace{-5pt}
}
\label{fig:mechinterp}
\end{figure*}
In Fig.~\ref{fig:failure}, we examine the learning dynamics for each failure mode. The figure indicates that the model initially acquires the skill to circumvent missteps (the blue line). Subsequently, it develops the ability to plan effectively, which is shown by a decrease in planning failures (the red line). By integrating these abilities---avoiding missteps and minimizing planning failures---the model is finally able to generate accurate paths for node pairs not seen during training.

\subsubsection{Mechanistic basis of the learned graph navigation algorithm}

Our results above elicit several intriguing behaviors attributable to stepwise inference.
We next take a more mechanistic lens to explain why these behaviors possibly occur.
We hypothesize that the model learns embeddings for the nodes of the graph that enable easy computation of an approximate distance measure.
This suggests that to move closer to the goal node, one can simply transition to the node that exhibits the least distance from the goal node. For the detailed intuition guiding our analysis, see Appendix~\ref{app:mechinterp}.

%we reconstruct a simple algorithm based on vector addition of the extracted value embeddings of the goal and the current node: compute an embedding by linearly adding the current node and goal node and output the next token whose token embedding has the highest inner product with this embedding (see Fig.~\ref{fig:mechinterp}b).
% 
To verify this, we first strip the model down to a single-head, self-attention layer. We visualize the attention scores for this minimal model in Fig.~\ref{fig:mechinterp}a, observing they are are concentrated on the goal node and the current node.
This suggests that the model utilizes only the embedding values of the goal $\red{X_{g}}$ and the current nodes $X_{\text{current}}$ to select the next token. 
Inspired by this observation, we develop a simplified algorithm that mimics the behavior of the  model, as outlined in detail in Fig.~\ref{fig:mechinterp}b.
First, we extract the value embeddings for $\textcolor{red}{X_g}$ and $X_{\text{current}}$ using the weight matrix $W_V$ from the self-attention layer, yielding $v_{g}$ and $v_{\text{current}}$, respectively. We then merge these embeddings into a single vector $v$, i.e., $v=v_{g}+v_{\text{current}}$. Finally, we determine the next token by identifying the node whose token embedding has the highest inner product with $v$. This operation mimics the logit computation in a full Transformer. 

In Fig.~\ref{fig:mechinterp}c, we demonstrate the simplified algorithm retrieved via the process above matches the accuracy of the full trained model. 
Furthermore, in Fig.~\ref{fig:mechinterp}d, we find that the paths generated by our simplified algorithm and those produced by the full trained model are nearly identical. Herein, we use a string edit distance metric~\citep{navarro2001guided} to quantify the similarity between the two sets of paths and find that over $75\%$ of paths are identical.

Given that accuracy is computed over test nodes not seen in the training data, it is likely that the model encodes a notion of distance between two nodes on the graph in its embedding, as we hypothesized. 
% In other words, the model learns a linear representation of the nodes within the graph, enabling navigation by identifying the node with the highest similarity (i.e., inner product) to the sum of the goal and current node embeddings. 
%
Indeed, in Fig.~\ref{fig:mechinterp}e, we find that the inner product of the embedding of $v_{g}$ with the token embeddings of $X_{\text{next}}$ is negatively correlated with the distance between these two nodes in the ground truth DAG; here, we used the average path length as a distance measure over the graph. 
% Using average path length as a proxy distance measure over the graph for now, we find that inner product of the embedding of $v_{g}$ with the token embeddings of $X_{\text{next}}$ is negatively correlated with the length of average path between these two nodes in the ground truth DAG (see  Fig.~\ref{fig:mechinterp}d).
% 
% 
Since potential nodes with shorter paths to the goal node have a higher logit value, this implies they will be more likely to be predicted, thus showing the origin of the short path bias we observed in Sec.~\ref{subsec:short_path_bias}. This is a mechanistic explanation of the pattern-matching behavior of ~\citet{dziri2023faith} in the context of our task.
% 
% Learning representations of states so that prediction is linear and corresponds to interpolation has been observed in LLMs at scale ~\citep{hosseini2023large}, in the brain ~\citep{henaff2019perceptual} and is an active area of research ~\citep{eysenbach2023contrastive}.
% 

%%%%%%%%%%%%%%%%%%%%%%%%%%%%%%%%%%%%%%%%%%%%%%%%%%%
% Few-shot CoT
%%%%%%%%%%%%%%%%%%%%%%%%%%%%%%%%%%%%%%%%%%%%%%%%%%%
\subsection{Navigation with exemplars}\label{sec:multi_graph}
The single graph setting let us explore \emph{zero-shot} navigation and stepwise reasoning, where the model relied on knowledge internalized over pretraining for stepwise navigation towards a goal. Next, we study how context can influence the model generated paths, how subgoals that are provided in-context can guide the model's navigation, and how the content of the exemplars affects the navigation path chosen by the model. 
Our results shed some light on and create hypotheses for (1) compositional generalization, (2) length generalization, and (3) impact of conflicting, long context.

\subsubsection{Compositional generalization} 
We find that the model can successfully follow the chain defined by the in-context exemplars. 
An example output produced by the model is in Fig.~\ref{fig:dpg_both}(b), highlighting the path the model takes through the chain of motifs $ g_3 \rightarrow g_4 \rightarrow g_2 \rightarrow g_9$. 
We also find that the model generalizes to arbitrary orders of motifs strung out, including those that did not occur consecutively in the training data, up to the length in the training data (see Fig.~\ref{fig:motif_length_generalization}). In other words, in-context control is capable of eliciting \textbf{\textit{compositional generalization}}~\citep{li2023dissectingcot}, if appropriately trained. Further, we see that the attentional patterns used by the model suggest that while navigating  across motifs, the model treats nodes across ghost edges as subgoals (see Appendix Fig.~\ref{fig:attention_pattern_motif}).

\subsubsection{Number of intermediate motifs}
\begin{figure}
\vspace{-10pt}
\centering
\includegraphics[width=0.85\linewidth]{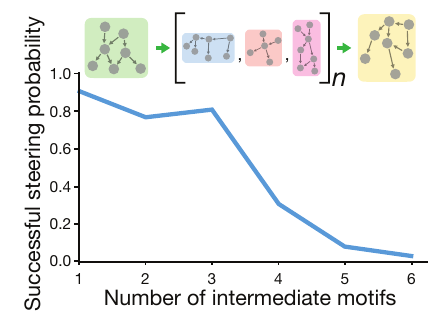}
\vspace{-10pt}
\caption{\textbf{In-context steerability and length generalization.} 
We vary the number of intermediate motifs $g_{\text{inter}}$ in a chain of motifs constructed for the particular context $ g_{i_1} \rightarrow \{ g_{\text{inter}} \rightarrow \}_n \rightarrow g_{i_K}$. The path generated by the model follows the path described by the chain in context until $n=4$, which is the maximum chain length in the training data. \vspace{-5pt}
\label{fig:motif_length_generalization}
}
\end{figure}

In Fig.~\ref{fig:motif_length_generalization}, we vary the number of exemplars provided to the model. This is equivalent to stringing together a longer chain of exemplar sequences across motifs to navigate over. 
We define successful steering via a product of indicator variables that measure (i) whether the path ended at the specified goal and (ii) that each ghost edge, and thus the intermediate motif, was present in the path. We computed the probability by averaging over distinct source nodes from $g_{i_1}$ and sink nodes from $g_{i_K}$.
We find that the model can generalize well to unseen orders of motifs up to the maximum number chained together in the training data, after which the model fails to navigate. 
We hypothesize that even when using stepwise inference methods at scale, \textit{the model will fail to generalize to reasoning chains longer than those present in its training data.} 

\subsubsection{Primacy bias towards the first exemplar in the case of conflict}
\begin{figure}
\vspace{-10pt}
\centering
\includegraphics[width=0.8\linewidth]{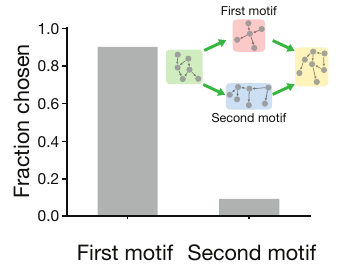}
\vspace{-10pt}
\caption{\label{fig:motif_conflict}\textbf{How does the model handle conflicting exemplars?} To construct the context, we selected an initial motif $g_{i_1}$, a terminal motif $g_{i_T}$ and two intermediate motifs $g_{\text{inter}}$ and $g'_{\text{inter}}$. We string them together so that the motif has two possible paths: $g_{i_1} \rightarrow g_{\text{inter}} \rightarrow g_{i_T}$ and $g_{i_1} \rightarrow g'_{\text{inter}} \rightarrow g_{i_T}$. In this case of two conflicting chains in-context, the model has a bias to pick the chain that appears first in context. }
\end{figure}

Language models are generally prompted with several exemplars in context. Some of these exemplars may have incorrect or even conflicting information with respect to other exemplars, for example in a multiple choice Q\&A task~\citep{hendrycks2020measuring, pal2022medmcqa, srivastava2022beyond}. 
The model has to choose the relevant information between these exemplars to solve the specified task.
Motivated by this, we quantitatively study the behavior of the model when a noisy context with exemplars with conflicting information are provided. Specifically, we study a case where two chains of motifs are used to design exemplars for our task, such that the exemplars start from the same set of initial and terminal motifs $g_{i_1}$ and $g_{i_T}$, but with distinct intermediate motifs $g_{\text{inter}}$ and $g'_{\text{inter}}$. 
The model is then prompted with $\blue{X_s} \in g_{i_1}$ and $\red{X_g} \in g_{i_T}$, after in-context exemplars in order: $e_{g_{i_1}, g_{\text{inter}}}, e_{g_{\text{inter}}, g_{i_T}}, e_{g_{i_2}, g'_{\text{inter}}}, e_{g'_{\text{inter}}, g_{i_T}}$. 
Results are shown in Fig.~\ref{fig:motif_conflict}. We find that the model does indeed navigate to the goal, thus following the prompt, but has a strong bias toward choosing a path defined by the first chain over the second, i.e., $g_{i_1} \rightarrow g_{\text{inter}} \rightarrow g_{i_T}$. 
This result is similar to what happens at scale with large context windows, where content in the middle of a long context window is ignored~\citep{liu2023lost}.

\section{Conclusion}
In this work, we introduced a synthetic graph navigation task to investigate the behavior, training dynamics, and mechanisms of Transformers under stepwise inference protocols. Despite its simplicity, our synthetic setup has provided key insights into the role of the structural properties of the data, a diversity-accuracy tradeoff in sampling, and a simplicity bias of stepwise inference protocols. In addition, we explored the model's navigation preferences and their controllability through in-context exemplars, modeled length generalization, and responses to longer contexts with conflicting exemplars.  Like all papers that rely on synthetic abstractions, our goal was to develop such hypotheses to explain an interesting phenomena seen in practical scenarios. 
A promising future direction for our work thus is to test the hypotheses we have formulated in large language models, as well as generalize and test the mechanistic interpretation of the learned Transformer algorithm in practical scenarios.

% We also discuss the limitations of our synthetic setup in Appendix~\ref{sec:limitations}.

% Our grounded synthetic task gives researchers control of variables that are generally not practical to control in the real-world. These ``knobs'' encompass a variety of parameters, such as the structure of the underlying graph (be it Bernoulli or hierarchical), the extent of training data paths, and inference temperature. Furthermore, in the in-context setting, we can manipulate the content, sequence, and volume of exemplars, among other factors. This framework stands as a unique playground or laboratory, though with its limitations, presenting insights into what stepwise inference is capable of and where its limitations lie.

\section*{Impact Statement}
This paper provides a comprehensive scientific analysis of a Transformer model that solves a small-scale synthetic task. We believe that the scientific findings presented in this paper will lay the groundwork for the development of more reliable and interpretable AI systems for the benefit of society.

% We believe that this work contributes to the better understanding of the inner-workings of the large-language models, 
% This paper presents work whose goal is to advance the field of Machine Learning. There are many potential societal consequences of our work, none which we feel must be specifically highlighted here.

\bibliography{icml2024}
\bibliographystyle{icml2024}

%%%%%%%%%%%%%%%%%%%%%%%%%%%%%%%%%%%%%%%%%%%%%%%%%%%%%%%%%%%%%%%%%%%%%%%%%%%%%%%
%%%%%%%%%%%%%%%%%%%%%%%%%%%%%%%%%%%%%%%%%%%%%%%%%%%%%%%%%%%%%%%%%%%%%%%%%%%%%%%
% APPENDIX
%%%%%%%%%%%%%%%%%%%%%%%%%%%%%%%%%%%%%%%%%%%%%%%%%%%%%%%%%%%%%%%%%%%%%%%%%%%%%%%
%%%%%%%%%%%%%%%%%%%%%%%%%%%%%%%%%%%%%%%%%%%%%%%%%%%%%%%%%%%%%%%%%%%%%%%%%%%%%%%

\appendix
\onecolumn

\section{Detailed Related Work}\label{sec:related_work}

\paragraph{Stepwise inference protocols}
Large language models (LLMs) have been shown to possess sophisticated and human-like reasoning and problem-solving abilities~\citep{srivastava2022beyond}.
Chain-of-thought or scratchpad reasoning refers to many similar and related phenomena involving multiple intermediate steps of reasoning \textit{generated internally and autoregressively} by the language model. 
First described by~\citet{nye2021show, kojima2022large}, adding prompts such as `\texttt{think step by step}' allows the LLM to autonomously generate intermediate steps of reasoning and computation, improving accuracy and quality of its responses. This is referred to as zero-shot chain-of-thought. A related set of phenomena, few-shot chain-of-thought prompting~\citep{wei2022chain} occurs when the language model is shown exemplars of reasoning before being prompted with a reasoning task. The model follows the structure of logic in these exemplars, solving the task with higher accuracy.  
Further, there have been several prompting strategies developed, all of which rely on sampling intermediate steps, such as tree-of-thoughts~\citep{yao2023tree}, graph-of-thoughts~\citep{besta2023graph}, program-of-thoughts~\citep{chen2022program}, self-ask~\citep{press2022measuring}. There are also methods which use more than one LLM, such as STaR~\citep{zelikman2022star}, RAP~\citep{hao2023reasoning}, Selection-Inference (SI)~\citep{creswell2022selection, creswell2022faithful}.

\paragraph{Understanding stepwise inference}
\citet{dziri2023faith} study how LLMs solve multi-step reasoning tasks and argue that models \textit{likely} fail because they reduce most multi-step reasoning tasks to linearized sub-graph matching, essentially learning `shortcut solutions'~\citep{liu2022Transformers}. \citet{momennejad2023evaluating} study in-context graph navigation in LLMs, finding that they fail to do precise planning. 
\citet{saparov2023language} introduce a synthetic dataset called PrOntoQA to systematically study the failure modes of chain of thought in the GPT3 family fine-tuned on the dataset and find that misleading steps of reasoning are a common cause of failure in the best-performing models.
~\citet{chen2023two} find that chain-of-thought fails at compositional generalization and counterfactual reasoning.
\citet{wang2022towards, schaeffer2023invalid} find that the content of the exemplars is less relevant to accuracy than their syntactic structure. \citet{razeghi2022impact} find that the accuracy of reasoning is correlated with the frequencies of occurrence in the pretraining dataset. 
Recently, a few works have used theoretical approaches to characterize and explain stepwise inference. \citet{li2023dissectingcot} study in-context learning of random MLPs and find that a Transformer that outputs the values of intermediate hidden layers achieves better generalization. \citet{feng2023towards} show that with stepwise reasoning, Transformers can solve dynamic programming problems, and~\citet{prystawski2023think} study reasoning traces in Transformers trained to learn the conditionals of a Bayes network.
There are also several puzzling phenomena in the prompts used to elicit few-shot chain-of-thought reasoning: chain-of-thought can be improved by sampling methods such as self-consistency~\citep{wang2022self}; prompts might not reflect the true reasoning process used by the language model, as identified by \citet{turpin2023language}; and the accuracy of the model can be sensitive to the order in which prompts are provided~\citep{lu2021fantastically}. 
% 
% \section{Limitations}\label{sec:limitations}
% In this work we have adopted the \emph{model-experimental systems approach}, an empirical strategy to precisely characterize and understand smaller, more steerable model systems with the ultimate goal of potentially transferring this understanding to larger-scale complex systems. It is important to clarify the trade-offs and limitations inherent in our approach. Drawing an analogy to the study of biological neural networks, where neural mechanisms identified in small-scale model organisms such as fruit flies or mice may not be directly applicable to medical applications involving the human brain, our observations should not be taken as definitive conclusions directly applicable to large-scale generative models. Instead, our study seeks to establish a minimal synthetic framework, identify data-centric control variables, and formulate mechanistic hypotheses. This lays the groundwork for more in-depth theoretical and empirical investigations of larger models.

\section{Why graph navigation?}\label{sec:why_graph}

\begin{figure}[]
\centering
\includegraphics[trim={0 0 0 0 pt}, clip, width=1.\columnwidth]{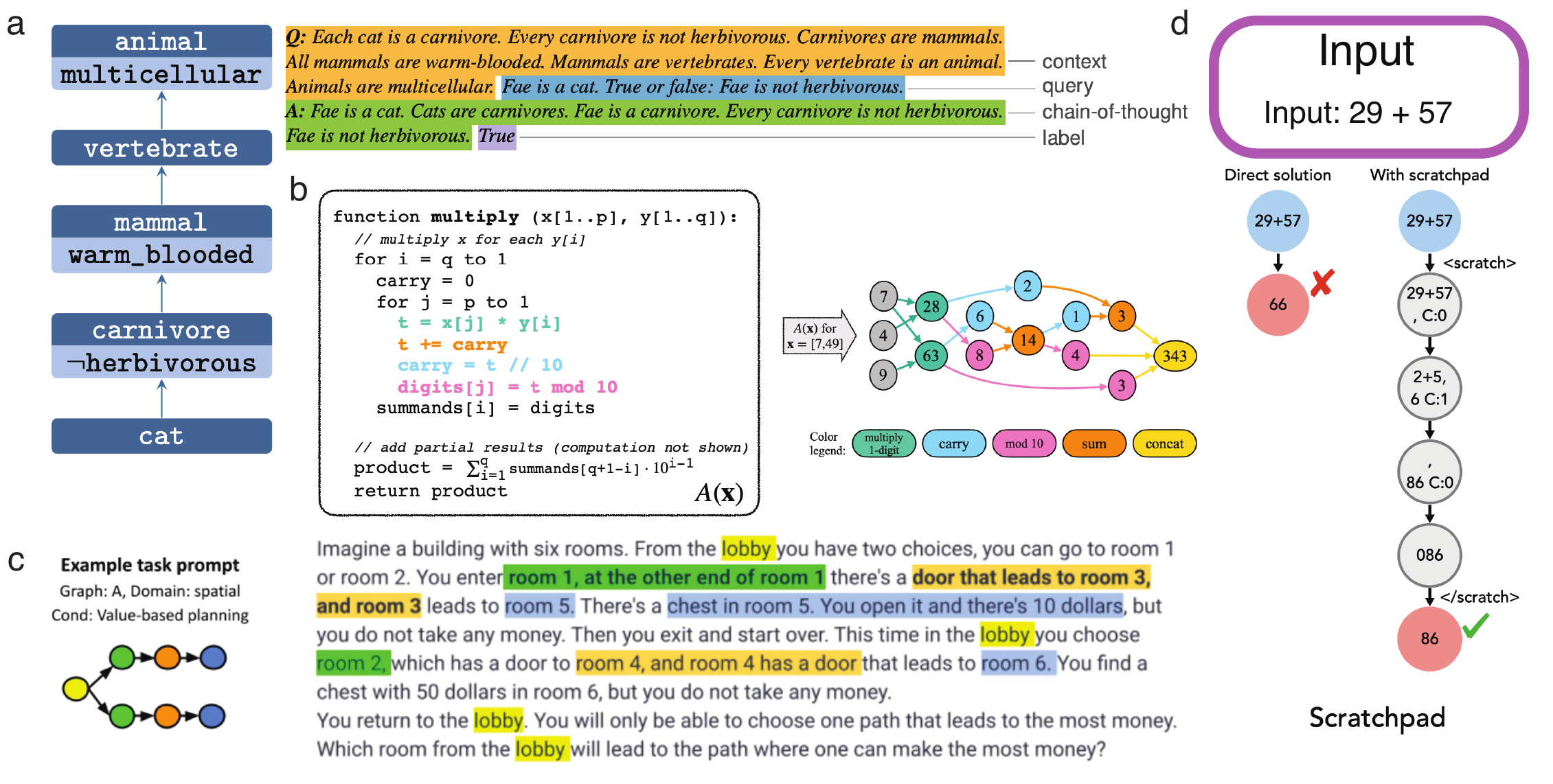}
\caption{
  \textbf{Examples of stepwise inference as graph navigation in LLM evaluations:} [Figures taken from respective papers] (a) An example graph created for a prompt (left) from the ProntoQ\&A dataset ~\citep{saparov2023language} (b) ~\citep{dziri2023faith} studies how simple algorithms such as multiplication of digits can be represented as a graph (c) CogEval~\citep{momennejad2023evaluating} studies many large scale LLMs such as ChatGPT-4 and Claude2 on planning and navigation tasks. (d) Mathematical expression evaluation in the case of additionof two numbers can be visualized as a series of steps of a digit-wise addition algorithm.
}
\label{fig:graph_navigation_sup} 
%\vspace{-10pt}
\end{figure}
In this section, we describe examples of various computational tasks that have been cast as graph navigation in literature to study Transformers and LLMs.

\begin{itemize}
    \item \textbf{First order logic: }~\citet{saparov2023language} study simple DAGs as models of first order logical reasoning. They construct \textit{ontologies} (see Fig.~\ref{fig:graph_navigation_sup}a) and prompt LLMs to do analogical reasoning. 
    \item \textbf{Mathematical expression evaluation: }~\citet{dziri2023faith} study mathematical expression evaluation in large scale LLMs as DAG navigation (see Fig.~\ref{fig:graph_navigation_sup}b). Any mathematical expression can be decomposed into elementary computations which are chained together.
    \item \textbf{Planning and spatial navigation: }~\citet{momennejad2023evaluating} evaluates many large scale LLMs such as ChatGPT-4 and Claude2 on synthetically designed planning and navigation tasks (see Fig.~\ref{fig:graph_navigation_sup}c).
    \item \textbf{Formal grammars and natural language: }~\citet{allen2023physics} studies Transformers trained on context-free grammars (CFGs) which are DAGs. Another motivation for the study of graph navigation comes from linguistics and natural language syntax~\citep{chomsky2002syntactic}. Every sentence in a language can broken down into its syntactic or parse tree, a special case of a directed acyclic graph. For example, the sentence `I drive a car to my college' can be parsed as the following graph: (`I': Noun phrase, `drive a car to my college': Verb Phrase) $\rightarrow$ (`drive': Verb, `a car': Noun Phrase, `to my college': Prepositional Phrase) $\rightarrow$ (`a': Determiner, `car': Noun), (`to': Preposition, `my college': Noun Phrase) $\rightarrow$ (`my': Determiner, `college': Noun).
\end{itemize}
Effective stepwise reasoning consists of several elementary logical steps put together in a goal-directed path that terminates at a precise state \cite{lavalle2006planning}. We argue that graph navigation problems provide such a fundamental framework for studying stepwise inference.
Graphs provide a universal language for modeling and solving complex problems across various domains. Whether it is optimizing network traffic, analyzing social networks, sequencing genetic data, or solving puzzles like the Travelling Salesman Problem, the underlying structure can often be mapped onto a graph ~\citep{cormen2022introduction, momennejad2023evaluating, dziri2023faith, saparov2023language}. 
% Inspired by algorithmic computational graphs and execution traces that capture how a program might reach a final state from an initial state by following elementary steps of computation, we model stepwise inference as traversal in a directed acyclic graph (DAG). 

\section{Setup and construction of graph and model}\label{sec:appendix_construction}

\subsection{Graph structures}\label{sec:graph_structures}
Here we describe the properties of the DAGs we use, the training setup, model architecture, and hyperparameters.
\begin{figure}[]
\centering
\includegraphics[trim={0 0 0 0 pt}, clip, width=0.7\columnwidth]{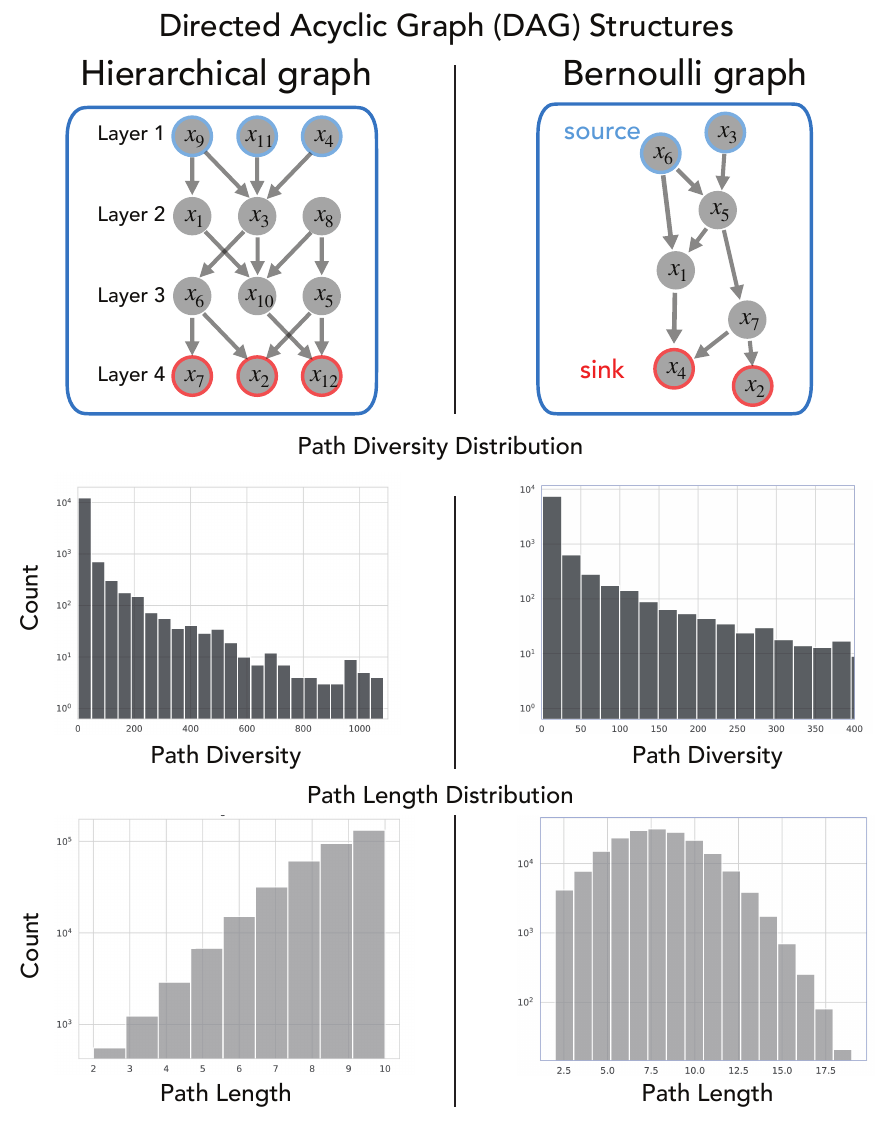}
\caption{
  \textbf{Construction and properties of Hierarchical and Bernoulli DAGs:} (top) Schematic of hierarchical and Bernoulli graphs. Hierarchical graphs are organized into layers with connections only between nodes of successive layers but Bernoulli graphs have no such structure. (middle) Path diversity is defined as the number of paths between any two path connected nodes. (bottom) Path length distributions: Owing to the hierarchical nature, the path length distribution is exponential in hierarchical graphs whereas it is more Gaussian-like for Bernoulli graphs.
}
\label{fig:path_properties}
%\vspace{-10pt}
\end{figure}

We use two DAG structures, as shown in Fig.~\ref{fig:path_properties}. Specifically, Bernoulli DAGs are constructed by randomly generating an upper-triangular matrix where each entry has probability $p$ of existing. Hierarchical DAGs are generated by predefining L sets of nodes and drawing an edge between a node $n_l$ in layer $l$ and $n_{l+1}$ in layer $l+1$ with probability $p$; we constrain the graph to be connected. These generation processes lead to different path diversity and path length distributions, which affect the efficacy of stepwise inference, as shown in our results. Below, we provide algorithms to generate our graph structures.

\begin{algorithm}
\caption{Generate Bernoulli connected DAG}
\begin{algorithmic}[1] % The number here denotes line numbering frequency, 1 for every line
\REQUIRE $\text{\texttt{numNodes}} > 0$, probability $p$ for edges
\STATE \texttt{nodeNames} $\gets$ [‘X’ + str(i) for i in range(\texttt{numNodes})]
\STATE \textbf{Function }CreateUpperTriangularMask(n, p)
    \STATE \texttt{matrix} $\gets$ random binary matrix with size $n \times n$ and probability $p$ for 1s
    \STATE \texttt{upperTriangular} $\gets$ extract upper triangular part of \texttt{matrix}
    \STATE \textbf{return} \texttt{upperTriangular}

\REPEAT
    \STATE \texttt{adjMatrix} $\gets$ CreateUpperTriangularMask(\texttt{numNodes}, p)
    \STATE \texttt{dag} $\gets$ create directed graph in NetworkX from \texttt{adjMatrix} and \texttt{nodeNames}
\UNTIL{\texttt{dag} is connected}
\end{algorithmic}
\end{algorithm}

\begin{algorithm}
\caption{Generate Hierarchical Connected DAG}
\begin{algorithmic}[1] % The number here denotes line numbering frequency, 1 for every line
\STATE $\texttt{p} \gets$ [probability of connection between layers]
\STATE $\texttt{nodesPerLayer} \gets$ [number of nodes in each layer]
\STATE $\texttt{numLayers} \gets$ [total number of layers]
\STATE $\texttt{numNodes} \gets \texttt{nodesPerLayer} \times \texttt{numLayers}$
\STATE \textbf{Function} CreateLayeredDAG$(\texttt{nodesPerLayer, numLayers, p})$
\STATE Initialize an empty directed graph $\texttt{G}$ in NetworkX
\FOR{$\texttt{currentLayer} \gets 1$ \textbf{to} $\texttt{numLayers} - 1$}
    \FOR{each node $j$ in $\texttt{currentLayer}$}
        \FOR{each node $k$ in $\texttt{currentLayer} + 1$}
            \IF{random number $\leq \texttt{p}$}
                \STATE Add edge from node $\texttt{X}_j$ to node $\texttt{X}_k$ in $\texttt{G}$
            \ENDIF
        \ENDFOR
    \ENDFOR
\ENDFOR
\STATE \textbf{return} $\texttt{G}$
\STATE \textbf{End Function}

\REPEAT
    \STATE $\texttt{dag} \gets$ CreateLayeredDAG$(\texttt{nodesPerLayer, numLayers, p})$
\UNTIL{$\texttt{dag}$ is connected}
\end{algorithmic}
\end{algorithm}

\newpage
\subsection{Motif construction} \label{Motif_construction}
In the multi-graph scenario, we first construct a set of $n$ graphs (in our experiments, we use Bernoulli DAGs with $n=10$) denoted by $G = \{g_1, g_2, ... , g_n\}$. To construct the training data, we first create all pairwise motif orders $\{(g_i \rightarrow g_j)\}$. For test evaluations, we held out $10$ out of these $45$ motif orders.

\subsubsection{Construction of exemplar sequences}

To provide examples in-context, we create exemplar sequences connecting motifs, say $g_{i_1}$ and $g_{i_2}$. In our construction, we select $\blue{X_{s}}$ to be \textit{source node} in $g_{i_1}$ and $\red{X_g}$ to be a \textit{sink node} in $g_{i_2}$. Further, we choose a \textit{sink} of $g_{i_1}$, $X_{\text{sink}}(g_{i_1})$ and a \textit{source} of $g_{i_2}$, $X_{\text{source}}(g_{i_2})$ and connect them via a \green{ghost edge}: $(X_{\text{sink}}(g_{i_1}), X_{\text{source}}(g_{i_2}))$. These intermediate nodes are \textbf{subgoals} for the path that the model has to produce. Finally putting everything together, the exemplar sequence has the following form: $ \red{\text{goal: }} \red{X_g} \, \blue{X_{s}} \dots X_{\text{sink}} \,(g_{i_1}) X_{\text{source}}(g_{i_2}) \dots \red{X_g}$. Here, $\blue{X_{s}} \dots X_{\text{sink}}$ is a path from a source to a sink in $g_{i_1}$ and $ X_{\text{source}}(g_{i_2}) \dots \red{X_g}$ is a path from a source to a sink in $g_{i_2}$. To be precise, we summarize this process into the algorithm below.

\begin{algorithm}
\caption{Generate In-context Exemplars}
\begin{algorithmic}[1] % The number here denotes line numbering frequency, 1 for every line
\REQUIRE $\{g_{i_1}, g_{i_2}\}$, two motifs across which a ghost edge will be placed.

    \STATE $\blue{X_s}$ $\gets$ \texttt{Sample}  sources($g_{i_1}$)
    \STATE $\red{X_g}$ $\gets$ \texttt{Sample}  sinks($g_{i_2}$)
    \STATE $\green{(X^{\text{ghost edge}}_{\text{pre}},X^{\text{ghost edge}}_{\text{post}})} \gets (\texttt{Sample}\text{ sinks}(g_{i_1}),\texttt{Sample}\text{ sources}(g_{i_2}))$
    \STATE $(\blue{X_s} \dots \green{X^{\text{ghost edge}}_{\text{pre}}})  \gets$ \texttt{Sample path} $(g_{i_1})$  
    \STATE $(\green{X^{\text{ghost edge}}_{\text{post}}} \dots \red{X_g})  \gets$ \texttt{Sample path} $(g_{i_2})$ 
    \STATE \textbf{return} $e_{g_{i_1}, g_{i_2}}$  $\gets \blue{X_s} \dots \green{X^{\text{ghost edge}}_{\text{pre}} X^{\text{ghost edge}}_{\text{post}}} \dots \red{X_g}$

\end{algorithmic}
\end{algorithm}

After providing a set of exemplar sequences in-context, we chain them together to create a longer sequence. To be precise, given a set of $K$ motifs $\{g_{i_1}, g_{i_2}, g_{i_3}, \dots g_{i_K} \}$, we have the set of $K-1$ ghost edges, one for each exemplar: $\{(X_{\text{sink}}(g_{i_1}), X_{\text{source}}(g_{i_2})), (X_{\text{sink}}(g_{i_2}), X_{\text{source}}(g_{i_3})), \dots (X_{\text{sink}}(g_{i_{K-1}}), X_{\text{source}}(g_{i_K})) \}$. To create the final path, we choose goal $\red{X_{g}} \in g_{i_1}$ and start $\blue{X_{s}} \in g_{i_K}$. This path has every ghost edge from the list present in it.

\subsection{Architecture details and loss function}
\label{sec:arch}

\subsubsection*{Loss function}

For training, we tokenize every node and we use the standard language modeling objective, next-token prediction with a cross entropy loss. Here $\text{target}_n$ is the 1-shifted version of the training sequence and $\mathbf{x}_n$ are the logit outputs of the model at the $n^{\text{th}}$ timestep.
\begin{equation}
    \mathcal{L}(\mathbf{x}_n, \text{target }n) = - \log \Big( \dfrac{\exp\small(\beta x_{n,\text{ target }n} \small)}{\sum_{v=0}^{\#\text{tokens}}\exp \small(\beta x_{n,v}\small)}\Big) = - \log \Big( \underbrace{\text{softmax}(\beta\mathbf{x}_{n})_{\text{target }n}}_{\text{prob}(\text{target } n)}\Big)
\end{equation}

\begin{table}[ht]
\centering
\begin{tabularx}{0.35\textwidth}{|l|X|}
\hline
\textbf{Hyperparameter}  & \textbf{Value} \\
\hline
 learning rate & $10^{-4}$ \\
\hline
Batch size &  64 \\
\hline
Context length & 32 \\
\hline
Optimizer  & Adam \\
\hline
Momentum & 0.9, 0.95 \\
\hline
Activation function  & GeLU \\
\hline
Number of blocks &  2\\
\hline
Embedding dimension  &  64\\
\hline
\end{tabularx}
\caption{Hyperparameters of the Transformer models used for all experiments except mechanistic analyses}
\label{table:hyperparameters}
\end{table}

For model architecture, we use a GPT based decode-only Transformer with a causal self-attention mask. Our implementation is based on the nanoGPT repository\footnote{available at \url{https://github.com/karpathy/nanoGPT}}.

\begin{figure}
\centering
\includegraphics[trim={0 0 0 0 pt}, clip, width=0.5\columnwidth]{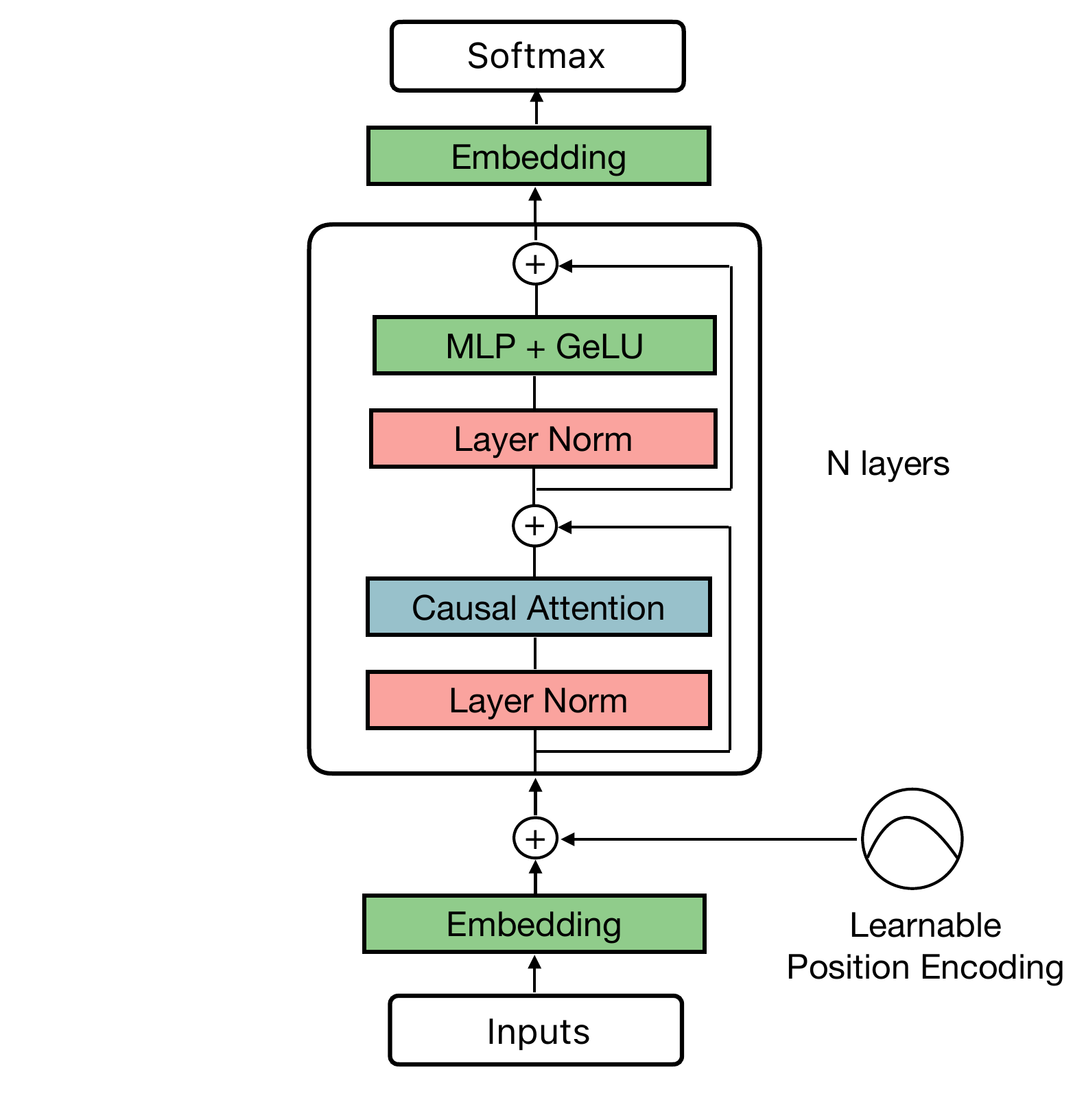}
\caption{
  \textbf{The architecture of GPT-style  ~\citep{radford2019language} decode-only Transformers.} Note the presence of both pre and post-LayerNorm in each Transformer block. Figure from methods section of~\citet{ramesh2023capable}.
}
\label{fig:Transformer_architecture}
\vspace{-10pt}
\end{figure}

The Transformer architecture consists of repeated blocks of pre-LayerNorm, causal multi-head self-attention, post-LayerNorm, and an MLP with skip connections (see Fig.~\ref{fig:Transformer_architecture}). The MLP contains two fully-connected layers with a GELU non-linearity~\citep{hendrycks2016gaussian}. The dimensionality of the hidden layer of the MLP is 4x the embedding dimensionality. We do not include any dropout in the model or biases in the linear layers. We use weight-tying~\citep{press2016using} in the embedding and un-embedding layers.  

To do the mechanistic study, we consider a 1 layer attention-only Transformer without a few modifications: We remove the MLP and post-LayerNorm and vary the embedding dimensionality from four to 64. This 1L Transformer is described by the following model equations. Here $\mathbf{
X_{\text{token}}} \in \mathcal{R}^{\text{vocab size} \times \text{{T}}}$ denotes the tokens, $W_E \in \mathcal{R}^{n_{\text{embd}}\times \text{vocab size}}$ is the positional embedding matrix, $W_{\text{pos}} \in \mathcal{R}^{n_{\text{embd}}\times T}$ is the token embedding matrix, and $\mathbf{X} \in \mathcal{R}^{n_\text{embd} \times \text{{T}}}$.

\begin{align*}
    \mathbf{X} &= W_E(\mathbf{X}_{\textbf{token}}) + W_{\text{pos}}(\mathbf{X}_{\textbf{token}})\\
    \mathbf{X} &= \text{LN}(\mathbf{X})\\
    \mathbf{X} &= \mathbf{X} + \text{softmax}(\mathbf{X}^T W_Q^TW_K\mathbf{X})W_V\mathbf{X}\\
    \mathbf{z} &= \text{softmax}(W_E^T\mathbf{X})\\
    \text{next token} &= \text{argmax}_{\text{all tokens}}\mathbf{z}
\end{align*}

\section{Training protocol for experiments}\label{sec:training_protocol}

For experiments in our setup without exemplars, we randomly generate either a hierarchical graph or a Bernoulli graph G with $N=200$ nodes. In the Bernoulli graph setting the probability of an edge $p=0.05$; similarly, in the hierarchical graph, the probability of an edge between a node in layer $l$ and layer $l+1$ is $p=0.05$. We choose 10 layers with 20 nodes each to match the number of nodes in the two graph types. We convert all the nodes to tokens, along with a special $\texttt{goal}$ token which corresponds to a $[\texttt{BOS}]$ token and an $\texttt{end}$ token which corresponds to an $[\texttt{EOS}]$ token. We use another token, $\texttt{pad}$, for padding as well.

\paragraph{Train-Test split} To generate training data corresponding to path connected node pairs, we first put all edges (which are paths of length one) into the training data. This procedure was done in all experiments to ensure that full knowledge of the graph was presented to the model.  Further, we generate all simple paths between every pair of nodes in the graph. A variable fraction of these paths are included in the training data, depending on experimental conditions which we outline below.

For experiments in Figs.~\ref{fig:stepwise_gap}a--b, we pick $20\%$ of the path-connected nodes and put all simple paths between them into the training data for each graph type. We also add an equal number of non-path connected nodes to the training data.

In Fig.~\ref{fig:stepwise_gap}c, for each value of the path-length threshold parameter, which sets the maximum length of paths in the training dataset, we pick paths corresponding to $20\%$ of the allowed path-connected node pairs and put them into the training data, while the remainder are held out evaluations. For the non-path connected pairs, we simply take all node pairs that are not path-connected and add a fraction of these node pairs into the training data, chosen to roughly balance the number of path-connected node pairs according to the experimental conditions. The rest are held-out for evaluation.

For the motif experiments in Fig.~\ref{fig:motif_length_generalization}, we generate a set of 10 motifs, each with a Bernoulli graph structure of 100 nodes with a bernoulli parameter $p=0.95$. We then divide the 45 possible motif orders into a set of 35 and 10 that we put into train and test respectively. For generating the context, we combine 3-6 motifs according to the allowed orders, and then sample exemplars as well as the final sequence that traverses the full motif chain by choosing start and goal nodes from the set of sources and sinks respectively. 

\section{Additional experimental results}

\begin{figure*}[ht!]
\includegraphics[width = \textwidth]{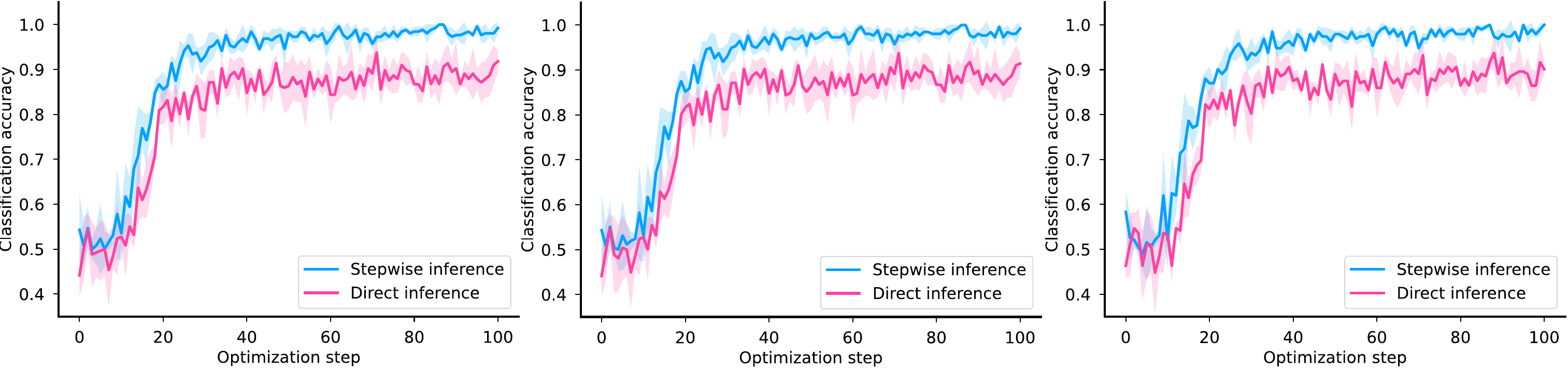}
\caption{
  \textbf{Persistence of stepwise inference gap with corrupted tokens:} In this experiment with setup identical to Fig.~\ref{fig:stepwise_gap}a-b, (a) $5\%$, (b) $10\%$ and (c) $20\%$ of tokens were randomly corrupted to mimic real world language data. The stepwise-inference gap persists.
}
\label{fig:noise_corrupted_gap}
\end{figure*}

\paragraph{Label noise in training data}
In Fig.~\ref{fig:noise_corrupted_gap}, we mimic real-world language data, abundant in ambiguity and polysemy, by corrupting (a) $5\%$, (b) $10\%$ and (c) $20\%$ of tokens in a single graph scenario. To achieve this, we replaced a randomly chosen $5\%$ and $10\%$ of the tokens in the training data with random tokens. We observe that the gap between stepwise inference and direct inference persists in both scenarios. This finding indicates that stepwise inference remains effective in more realistic settings with noise. 

\begin{figure}
\centering
\includegraphics[width=\linewidth]{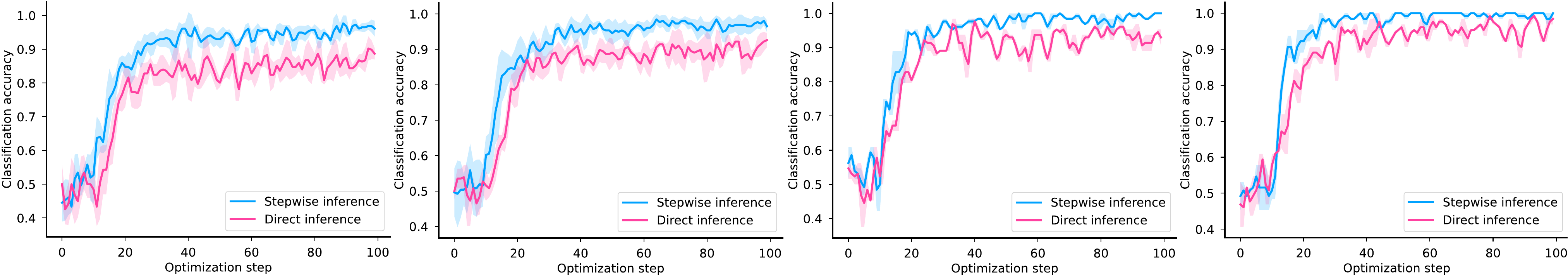}

\caption{
   \textbf{The effect of varying edge density in the single graph scenario:} Here we vary $\texttt{p}$, the edge density of connectivity in the graph, from $0.08$ in the left-most plot to $0.12$ in the right-most plot, in steps of $0.01$. The stepwise inference gap persists in all cases.}
\label{fig:training_curve_p}
\end{figure}

\paragraph{Varying edge density}
In Fig.~\ref{fig:training_curve_p}, we swept the density of the graph from 0.08 to 0.12 on a hierarchical graph. We observe a stepwise inference gap in all cases. The stepwise inference gap becomes smaller for larger densities. This is because the more likely the nodes are to be connected, the more likely it is for shortest paths to exist between nodes and thus less ``stitching" is needed~\citep{broadbent1957percolation}.

\begin{figure}
\centering
\includegraphics[width=0.3\linewidth]{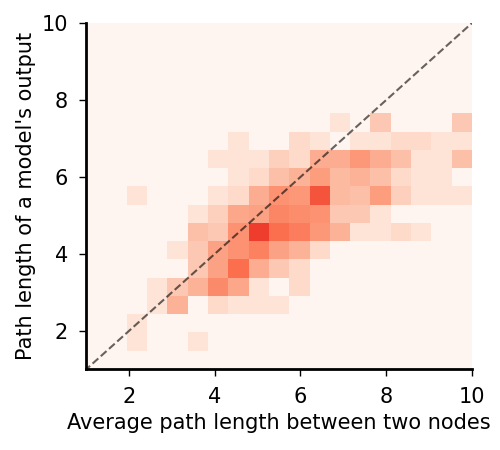}
\caption{\textbf{Model outputs are biased toward shorter paths.}}
\label{fig:shortpath_bias_density}
\end{figure}

\paragraph{Short path bias}
Fig.~\ref{fig:shortpath_bias_density} presents a density plot comparing the average lengths of actual paths with those generated by the model in a Bernoulli graph. This observation verifies the model tends to produce shorter paths between a given pair of start and goal nodes.

\begin{figure}
\centering
\includegraphics[width=\linewidth]{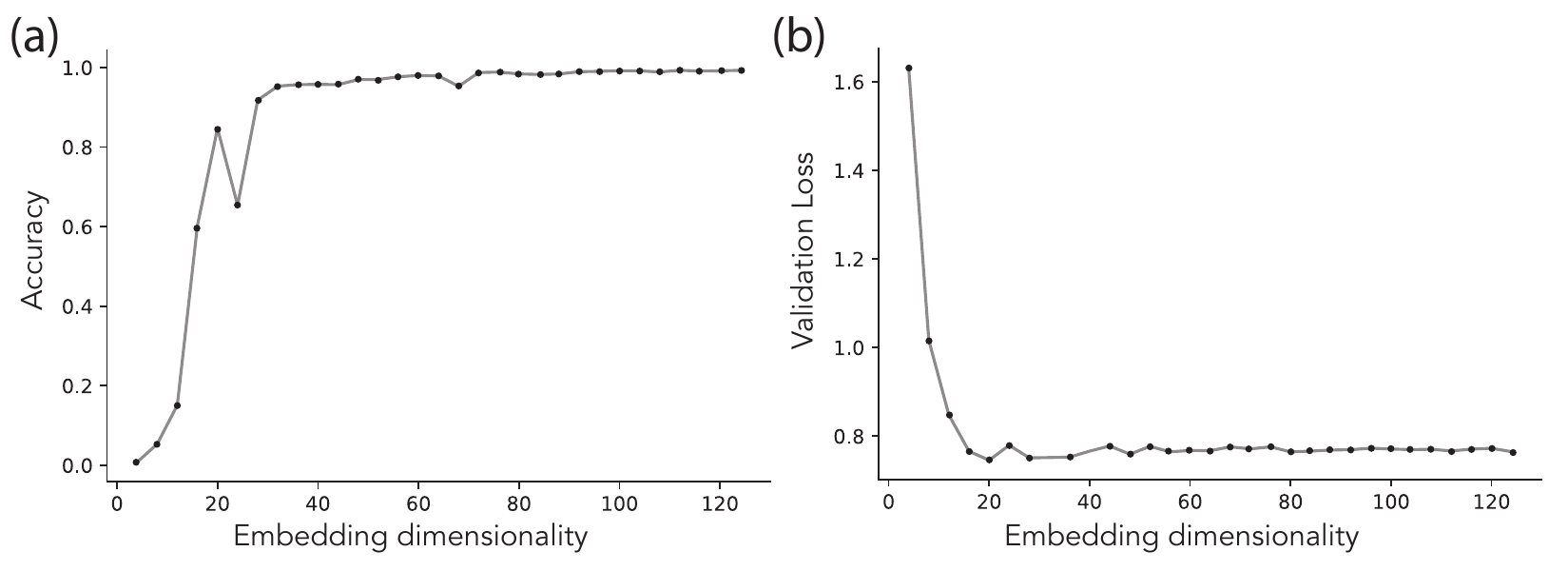}
\caption{\textbf{Varying embedding dimensionality in 1 layer models:} We find that there is a critical embedding dimensionality (around 20 for a Bernoulli graph of size 200 nodes and $p=0.05$) above which the model can generalize to all held-out node pairs.}
\label{fig:embedding_dimensionality}
\end{figure}

\paragraph{Effect of varying embedding dimensionality in the single graph scenario}
Here we consider the 1-layer Transformer without MLP and post-LayerNorm and ask the following question: for a fixed underlying graph size and training data, how does the model performance vary as we sweep embedding dimensionality. Intuitively, if the embedding dimensionality is large, the model should be able to generalize better by learning a better embedding of the node tokens. We see that beyond a critical dimensionality (which is around $20$ for a graph of 200 nodes), the model generalizes to all held out (start, goal) node pairs with a fairly abrupt transition (see Fig.~\ref{fig:embedding_dimensionality}). 

\section{Intuition guiding the mechanistic analysis}
\label{app:mechinterp}
In this section, we present the intuition that served as the hypothesis guiding our mechanistic analysis. 

Consider the optimal maximum likelihood estimator designed to solve our graph navigation task. 
Given a start node $\blue{X_{s}}$ and an incomplete sequence of predicted nodes $X_1, \dots, X_{k}$ in the pursuit of navigating to  the goal node $\red{X_{g}}$, the estimator works the following way:
\begin{equation*}
    X_{\text{next}} = \arg\max_{X'} P(X'| \blue{X_{s}}; X_1, \dots, X_{k}; \red{X_{g}})
\end{equation*}
Since the task is conditionally Markovian: the choice of the next step will be independent of the history when conditioned on $\red{X_g}$ and $X_{k}$. Accordingly, we have:
\begin{align*}
    X_{\text{next}} &= \text{argmax}_{X'} P(X'| X_{k}, \red{X_{g}})\\
    &= \arg\max_{X'} \dfrac{P(\red{X_{g}} | X', X_{k}) P(X', X_{k})}{P(\red{X_{g}}, X_{k})}
\end{align*}

This decomposition leads to interpretable terms which shed light on what algorithm the model might use:
\begin{equation*}
\begin{split}
    \log P(X'| X_{k}, \red{X_{g}}) = \log P(\red{X_{g}}|X', X_{k}) + \log P(X', X_{k}) - \log P(\red{X_{g}}, X_{k})
\end{split}
\end{equation*}

These terms can be interpreted as follows:
\begin{enumerate}
    \item $\log P(\red{X_{g}}, X_{k})$ describes the prompt.
    \item $\log P(X', X_{k})$ describes the knowledge of the \textbf{world model}: How well does the model know the ground truth structure of the graph?
    \item $\log P(\red{X_{g}}|X', X_{k})$ corresponds to \textbf{goal-directed behavior}: What $X'$ is most likely to lead to the goal?
\end{enumerate}

Let $\mathcal{C}(X_k)$ denote the subset of nodes in the graph that are \textit{children} of the node $X_k$. Then, while selecting the next token that has the highest likelihood, note that terms (1) and (2) cannot be optimized over: the former does not depend on $X'$ and the latter, for the optimal predictor, will be $\nicefrac{1}{|\mathcal{C}(X_k)|}$ if $X' \in \mathcal{C}(X_k)$ and $0$ otherwise.
Accordingly, the only term that can be optimized over is the third one, i.e., the one that measures how likely the goal is if the next state is $X'$. However, due to term (2), $X' \in \mathcal{C}(X_k)$---that is, the possible set of next tokens is constrained to the set $\mathcal{C}(X_k)$.

Now, exploiting the task's conditional Markovian nature again, we have $P(\red{X_{g}}|X', X_{k}) = P(\red{X_{g}}| X': X' \in \mathcal{C}(X_k))$. 
% Here $X'$ is restricted to the set of all nodes $\in \text{children}(X_k)$
% 
Heuristically, assume that $P(\red{X_{g}}|X') \propto \text{e}^{- d(\red{X_{g}}, X')} \cdot \mathbb{I}(X' \in \mathcal{C}(X_k))$, where $d(X_i, X_j)$ is a measure that describes the distance between nodes $X_i$ and $X_j$, while respecting the topology of the graph, and $\mathbb{I}$ is an indicator function that is $1$ if its input is True, and $0$ if not.
Then, we have $\log P(\red{X_{g}}|X', X_{k}) \propto - d(\red{X_{g}}, X') \cdot \mathbb{I}(X' \in \mathcal{C}(X_k))$. 

The intuitive argument above, though likely to be approximate, suggests that a possible solution the model can learn via autoregressive training in our graph navigation setup is (i) compute the distance between all neighboring nodes of the current node and the goal node, (ii) move to the node that has the least distance, and (iii) repeat.
The algorithm we uncover in our analysis in Sec.~\ref{sec:mechinterp} in fact functions in a similar way: the model is constantly computing a inner product between the goal token's representation and the embeddings of all tokens; we find this inner product is highest for the neighbors of the current token. 
Then, the highest inner product token is outputted and the process is repeated.
Since the embeddings are not normalized, this inner product is not exactly the Euclidean distance---we expected as much, since the topology of the graph will have to be accounted for and learning an inner product based metric will be easier for a model (because most operations therein are inner products).

\begin{figure}[t]
\centering
\includegraphics[trim={0 0 0 0 pt}, clip, width=0.5\columnwidth]{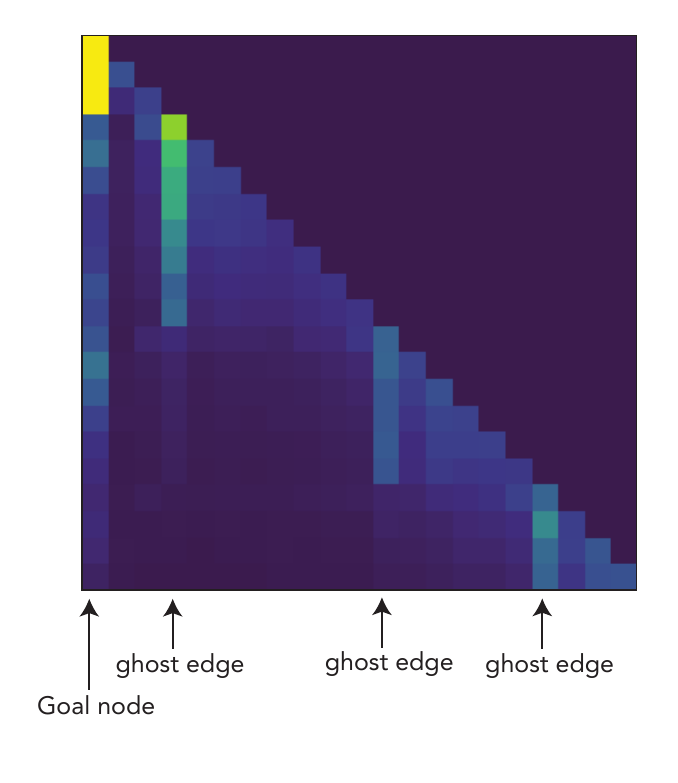}
\caption{
  \textbf{Attention pattern after in-context exemplars:} Here we visualize the portion of the attention map after prompting with four in-context exemplar sequences. The model generates attentional patterns that treat the node across ghost edges as a subgoal. 
}
\label{fig:attention_pattern_motif}
\vspace{-10pt}
\end{figure}

Further, in the case of motifs, we expect that the model contructs a path through checkpoints defined by ghost edges, which act as subgoals. To explain, given a set of K motifs strung together in-context $\{g_{i_1}, g_{i_2}, g_{i_3}, \dots g_{i_K} \}$, we have the set of K-1 ghost edges, one for each exemplar: $\{(X_{\text{sink}}(g_{i_1}), X_{\text{source}}(g_{i_2})), (X_{\text{sink}}(g_{i_2}), X_{\text{source}}(g_{i_3})), \dots (X_{\text{sink}}(g_{i_{K-1}}), X_{\text{source}}(g_{i_K})) \}$. Thus, we hypothesize that the model identifies all K-1 ghost edges from its context and plans sub-paths to each ghost edge in pursuit of the goal. Preliminary analyses of attention patterns in Fig.~\ref{fig:attention_pattern_motif} provides evidence for this hypothesis.

\subsection{Generalizing static word embeddings to 3-way relations}
Static word embedding algorithms such as Word2vec are trained by sampling pairs of words $(w_i,w_c)$ that appear in the same context and adjusting their embeddings so that their inner product is higher than an inner product with the embedding of word $w_i$ and a randomly sampled word from the vocabulary. 
The algorithm can be understood as performing a low-rank factorization of the matrix of co-occurrence statistics. In the case of Word2vec, the matrix is factorized as $P = I \cdot C$ where $I$ contains \emph{word vectors} in its rows and $C$ contains \emph{word covectors} in its columns. Therefore, every word has two types of embeddings. One is used when the word appears in the first position in the pair (which corresponds to center words), and the second when it appears in the second position (which corresponds to context words). 

Inspired by the interpretability results, we argue our graph navigation task can also be solved using a similar low-rank factorization method that is generalized to 3-way relations. In this case, the tensor $T$ to be factorized is third-order, and for each node, we have three types of embeddings: one which is used when the node acts as a goal, one when it acts as the current state, and one when it acts as the next possible state. 

Since we do not deal with a natural corpus with different frequency of occurrence of individual nodes, we can we set the numbers $T_{ijk}$ to be proportional to the length $l_{ijk}$ of a path which goes through an ordered pair of neighbour nodes $(i,j)$ to a node $k$. If there is no such path, we set the length to infinity. The target value $T_{ijk}$ can be seen as a preference for a node $j$ when the goal is to reach the node $k$ from the node $i$; it is equal to ${l_{ijk}^{-1}}/{\sum_{j'}{l_{ij'k}^{-1}}}$.

Inspired by the learned algorithm, we can use low-rank tensor factorization to approximate this matrix by the following expression $T_{ijk} \approx \hat{T}_{ijk}= (\mathbf{s}_i + \mathbf{g}_k) \cdot \mathbf{n_i}$, where $\mathbf{s_i}$, $\mathbf{g}_k$, $\mathbf{n_i}$ are the three types of learnable embeddings. Therefore, by interpreting the trained Transformer, we can obtain a simple algorithm that can be potentially useful in setups that deal with 3-way relationships. We leave this for future work.

%When trained on the same graph as used for the main experiments, this method quickly reaches the same accuracy. 

\end{document}